%% file: acl_latex.tex
\pdfoutput=1
\documentclass[11pt]{article}

\usepackage{acl}

\usepackage{times}
\usepackage{latexsym}
\usepackage[T1]{fontenc}
\usepackage[utf8]{inputenc}
\usepackage{microtype}
\usepackage{inconsolata}

\usepackage{graphicx}
\usepackage[table]{xcolor}
\usepackage{amsmath, amssymb}
\usepackage{wrapfig}
\usepackage{multirow}
\usepackage{booktabs}            
\usepackage{colortbl}
\usepackage{xcolor} 
\usepackage[section]{placeins}   
\usepackage{dblfloatfix}         
\usepackage{balance}             
\usepackage{caption}
\usepackage{tcolorbox}           

\usepackage{soul}
\setul{0.3ex}{0.12ex}            
\setuldepth{ConStory-Bench}      
\newcommand{\yuhao}[1]{{\color{violet}#1}} 
\makeatletter
\def\fps@figure{!t}
\def\fps@table{!t}
\makeatother

\definecolor{mplBlue}{HTML}{1f77b4}
\definecolor{mplOrange}{HTML}{ff7f0e}
\definecolor{mplGreen}{HTML}{2ca02c}
\definecolor{mplRed}{HTML}{d62728}
\definecolor{mplPurple}{HTML}{9467bd}

\title{Lost in Stories: Consistency Bugs in Long Story Generation by LLMs}
\author{
  Junjie Li\textsuperscript{1} \quad
  Xinrui Guo\textsuperscript{1} \quad
  Yuhao Wu\textsuperscript{2} \quad
  Roy Ka-Wei Lee\textsuperscript{2} \quad
  Hongzhi Li\textsuperscript{1} \quad
  Yutao Xie\textsuperscript{1} \\[6pt]
  \textsuperscript{1}Microsoft, Beijing, China \\
  \textsuperscript{2}Singapore University of Technology and Design \\[4pt]
  \texttt{lij850601@gmail.com} \quad
  \texttt{xingu@microsoft.com} \quad
  \texttt{wu\_yuhao@mymail.sutd.edu.sg}
}

\begin{document}
\maketitle

\begin{abstract}

What happens when a storyteller forgets its own story? Large Language Models (LLMs) can now generate narratives spanning tens of thousands of words, but they often fail to maintain consistency throughout. When generating long-form narratives, these models can contradict their own established facts, character traits, and world rules. Existing story generation benchmarks focus mainly on plot quality and fluency, leaving consistency errors largely unexplored. To address this gap, we present \textbf{ConStory-Bench}, a benchmark designed to evaluate narrative consistency in long-form story generation. It contains 2,000 prompts across four task scenarios and defines a taxonomy of five error categories with 19 fine-grained subtypes. We also develop \textsc{ConStory-Checker}, an automated pipeline that detects contradictions and grounds each judgment in explicit textual evidence. Evaluating a range of LLMs through five research questions, we find that consistency errors show clear tendencies: they are most common in factual and temporal dimensions, tend to appear around the middle of narratives, occur in text segments with higher token-level entropy, and certain error types tend to co-occur. These findings can inform future efforts to improve consistency in long-form narrative generation.
Our project page is available at \url{https://picrew.github.io/constory-bench.github.io/}.

\end{abstract}

\input{sec/1_change_intro.tex}
\input{sec/3_Constory_benchmark.tex}
\input{sec/4_change_evalution.tex}
\input{sec/2_related.tex}

\input{sec/5_conclusion.tex}
\input{sec/6_limitations.tex}

\FloatBarrier
\bibliography{custom}

\clearpage
\input{sec/appendix.tex}

\end{document}

%% file: sec/1_change_intro.tex
\section{Introduction}

Long-form narrative generation has become a key capability for large language models (LLMs) empowering a wide range of applications including, e.g. content creation, storytelling, and educational authoring.
As context windows expand, models must maintain \emph{consistency} across thousands of tokens by accurately tracking entities and events, preserving world rules, and sustaining coherent stylistic conventions, rather than merely producing locally fluent text. 
Recent research has advanced long-context understanding and long-form generation capabilities, yet these efforts have not systematically isolated cross-context contradictions or provided reproducible evaluation mechanisms at scale~\cite{bai2024longbench,bai2025longbench,an2024eval,wu2024longgenbench,que2024hellobench}. 
Within narrative generation, existing planning-based approaches~\cite{zhou2023recurrentgpt,wang2023improving,xie2024creating,gurung2024chiron,wen2023grove} and creative writing evaluations~\cite{ismayilzada2024evaluating,xie2023next,wang2024weaver} focus primarily on plot coherence and fluency, leaving global consistency underexplored.
Furthermore, while LLM-as-a-judge protocols show promise for automated evaluation, existing approaches typically lack explicit textual evidence and interpretable rationales~\cite{lee2024checkeval,pereira2024check,tan2024proxyqa,chen2024humans,zheng2023judging}.

\begin{figure*}[t]
    \centering
    \includegraphics[width=1\linewidth]{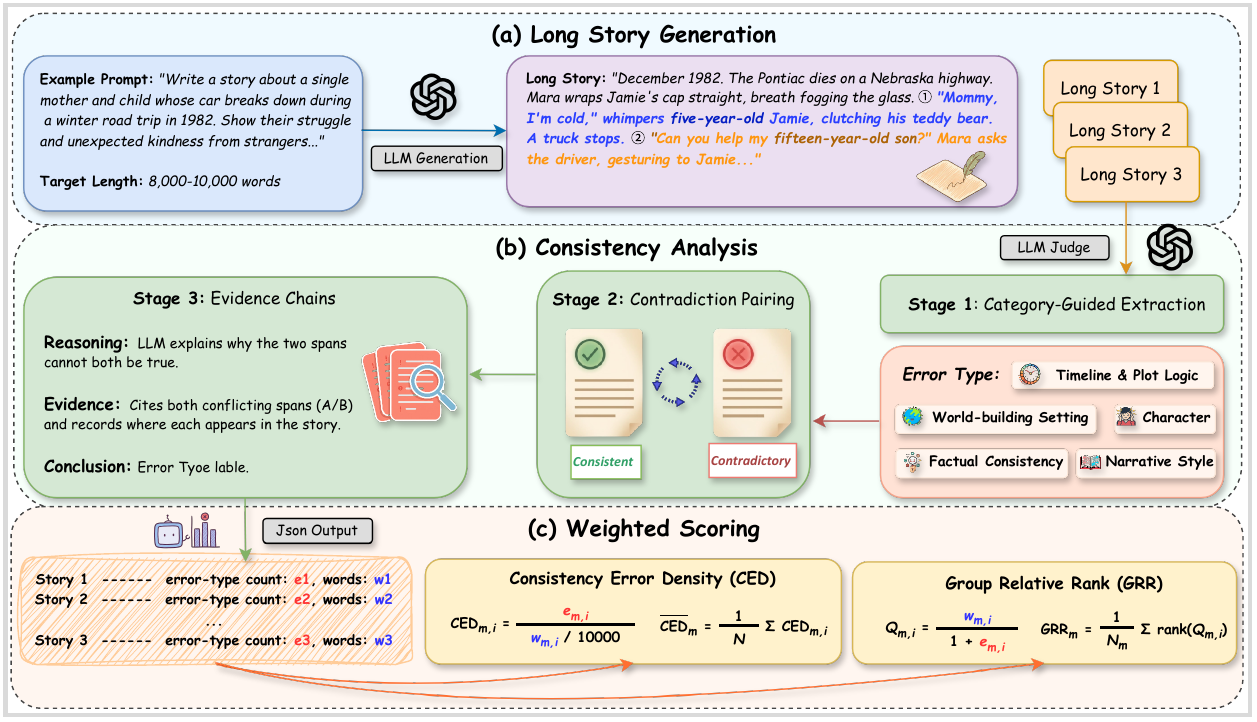}
    \caption{\textbf{Overview of ConStory-Bench.} The framework comprises three components: (a) a 2{,}000-prompt benchmark for long story generation (Targeting 8,000--10,000 words), (b) \textsc{ConStory-Checker}, a three-stage pipeline that extracts errors across five categories, pairs contradictions, and constructs evidence chains, and (c) standardized scoring via Consistency Error Density (CED) and Group Relative Rank (GRR).}
    \label{fig:overview}
\end{figure*}

To fill this gap, we present \textbf{ConStory-Bench}, a benchmark for evaluating narrative consistency in long-form story generation.
We also develop \textsc{ConStory-Checker}, an automated evaluation pipeline that detects contradictions and grounds each judgment in explicit textual evidence with exact quotations.
ConStory-Bench comprises 2{,}000 prompts across four narrative task scenarios and defines a five-dimension taxonomy with 19 fine-grained error subtypes.
An overview of the benchmark and pipeline is provided in Figure~\ref{fig:overview}.

We structure our investigation around the following \textbf{Research Questions}:
\textbf{(1)} \textit{To what extent do current LLMs maintain narrative coherence in ultra-long text generation, and do different models exhibit similar distributions of consistency error types?}
\textbf{(2)} \textit{How do consistency errors scale as a function of output length across different LLM architectures?}
\textbf{(3)} \textit{What underlying factors contribute to the emergence of consistency errors, and are there identifiable signals that reliably predict their occurrence?}
\textbf{(4)} \textit{Do different types of consistency errors systematically co-occur, or do they arise independently?}
\textbf{(5)} \textit{How are consistency errors distributed across positions within long-form generated narratives?}

\yuhao{Our main contributions are as follows:}
\begin{itemize}
\item We introduce \textbf{ConStory-Bench}, a benchmark for evaluating narrative consistency in long-form story generation, with four task scenarios and a taxonomy of five error categories and 19 fine-grained subtypes.
\item We develop \textsc{ConStory-Checker}, an automated evaluation pipeline that detects contradictions and supports each judgment with exact textual evidence.
\item We present evaluation results for a broad range of text generation systems, spanning proprietary and open-source models, capability-enhanced models, and agentic generation systems, and conduct a systematic analysis guided by five research questions.
\end{itemize}


%% file: sec/3_Constory_benchmark.tex
\section{ConStory-Bench}
\label{sec:constory-bench}

We present ConStory-Bench, a benchmark for evaluating consistency in long-form narrative generation. The benchmark uses an LLM-as-judge pipeline to detect consistency errors and classify them into fine-grained categories. Section~\ref{sec:construction} describes the data collection and prompt construction procedure, Section~\ref{sec:error-taxonomy} introduces the error taxonomy, and Section~\ref{sec:detection-pipeline} presents the automated evaluation pipeline.

\begin{figure}[t!]
\centering
\includegraphics[width=\columnwidth]{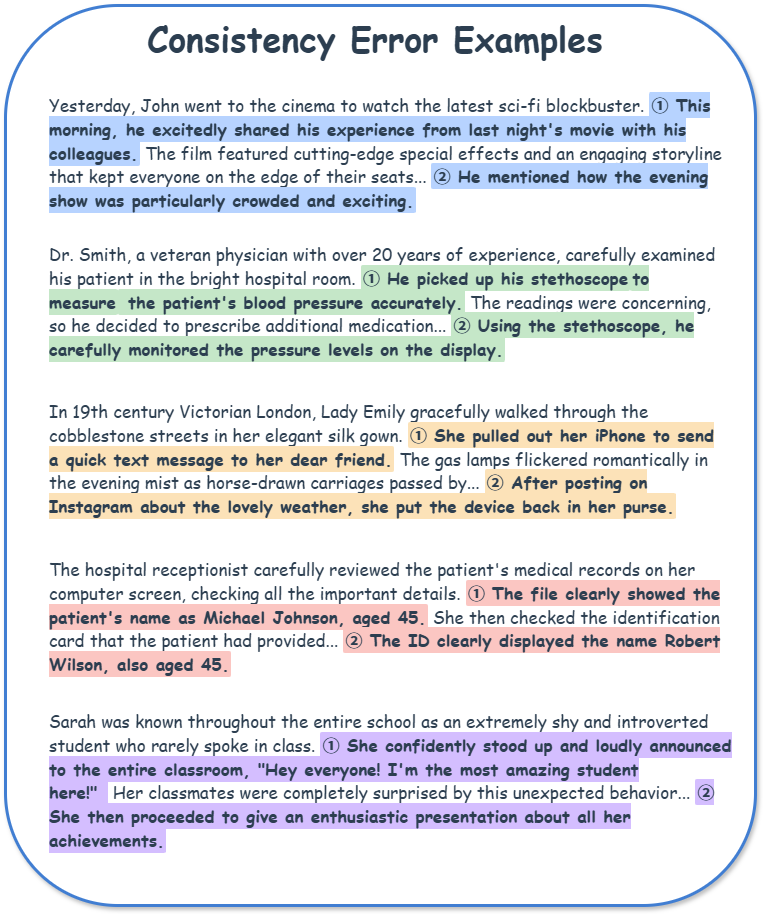}
\caption{Representative consistency error examples sampled from real LLM-generated stories on ConStory-Bench. Highlighted segments show contradictions in \textcolor[RGB]{41,128,185}{Timeline \& Plot Logic}, \textcolor[RGB]{39,174,96}{Characterization}, \textcolor[RGB]{241,196,15}{World-building \& Setting}, \textcolor[RGB]{230,126,34}{Factual \& Detail Consistency}, and \textcolor[RGB]{155,89,182}{Narrative \& Style}.}
\label{fig:consistency-error-examples}
\end{figure}

\subsection{Dataset Construction}
\label{sec:construction}

\paragraph{Sources and Selection.}
We collect seed stories from seven diverse public corpora: \textsc{LongBench}~\cite{bai2024longbench}, \textsc{LongBench\_Write}~\cite{bai2024longwriter}, \textsc{LongLamp}~\cite{kumar2024longlamp}, \textsc{TellMeAStory}~\cite{akoury2020storium}, \textsc{WritingBench}~\cite{wu2025writingbench}, \textsc{WritingPrompts}~\cite{fan2018hierarchical}, and \textsc{WikiPlots}~\cite{riedl-2017-wikiplots}. We extract both creative writing queries and full-length narratives from these corpora.

\paragraph{Prompt Construction via LLM Rewriting.}
We convert the collected stories into task-specific prompts to elicit long-form narrative generation from models. For each story, we first assign one of four task types based on its narrative structure and content: \emph{generation} - produce a free-form narrative given only a minimal plot setup, \emph{continuation} - extend an initial story fragment into a complete, coherent narrative, \emph{expansion} - develop a long-form story from a concise yet relatively complete plot outline by elaborating implicit details and events, \emph{completion} - write a full story with predefined beginning and ending, given minimal guidance for the intervening plot. Using \texttt{o4-mini}, we then rewrite each story into a prompt tailored to its assigned task type, grounding prompts in authentic narrative elements from the source stories while constraining target generation length to 8,000--10,000 words. Finally, we perform quality control through: (i)~MinHash-based deduplication to remove near-duplicate prompts, and (ii)~filtering low-quality or trivial cases through manual inspection and automated heuristics. This process yields 2{,}000 high-quality prompts distributed across the four task types (Table~\ref{tab:dataset-statistics}). Detailed task specifications and representative prompt examples are provided in Appendix~\ref{sec:benchmark-construction}.

\begin{table}[t!]
\centering
\small
\begingroup
\renewcommand{\arraystretch}{1.35}
\begin{tabular}{lcc}
\toprule
\textbf{Task Type} & \textbf{Count} & \textbf{Percentage} \\
\midrule
Generation   & 751  & 37.5\% \\
Continuation & 432  & 21.6\% \\
Expansion    & 422  & 21.1\% \\
Completion   & 395  & 19.8\% \\
\midrule
\textbf{Total} & \textbf{2,000} & \textbf{100\%} \\
\bottomrule
\end{tabular}
\endgroup
\caption{Statistics of ConStory-Bench across four task types.}
\label{tab:dataset-statistics}
\end{table}

\subsection{Consistency Error Taxonomy}
\label{sec:error-taxonomy}

To enable systematic evaluation, we develop a hierarchical taxonomy grounded in narrative theory and prior research on story understanding~\citep{ismayilzada2024evaluating,xie2023next}. 
The taxonomy comprises five top-level categories and 19 fine-grained error types (Table~\ref{tab:error-taxonomy}), encompassing contradictions that emerge across temporal logic, character memory, world-building rules, factual details, and narrative style. 
Representative error cases with detailed annotations are presented in Figure~\ref{fig:consistency-error-examples}.

\begin{table}[t!]
\centering
\small
\setlength{\tabcolsep}{8pt}
\renewcommand{\arraystretch}{1.2}
\begin{tabular}{p{3cm}@{\hspace{1em}}l}
\toprule
\textbf{Error Type} & \textbf{Sub Error Type} \\
\midrule
 \rowcolor{blue!10}
\textbf{Timeline \& Plot} & Absolute Time Contradictions \\
 \rowcolor{blue!10}
\textbf{Logic} & Duration Contradictions \\
 \rowcolor{blue!10}
 & Simultaneity Contradictions \\
 \rowcolor{blue!10}
 & Causeless Effects \\
 \rowcolor{blue!10}
 & Causal Logic Violations \\
 \rowcolor{blue!10}
 & Abandoned Plot Elements \\
 \rowcolor{green!10}
\textbf{Characterization} & Memory Contradictions \\
 \rowcolor{green!10}
 & Knowledge Contradictions \\
 \rowcolor{green!10}
 & Skill Fluctuations \\
 \rowcolor{green!10}
 & Forgotten Abilities \\
 \rowcolor{yellow!10}
\textbf{World-building \&} & Core Rules Violations \\
 \rowcolor{yellow!10}
\textbf{Setting} & Social Norms Violations \\
 \rowcolor{yellow!10}
 & Geographical Contradictions \\
 \rowcolor{orange!10}
\textbf{Factual \& Detail} & Appearance Mismatches \\
 \rowcolor{orange!10}
\textbf{Consistency} & Nomenclature Confusions \\
 \rowcolor{orange!10}
 & Quantitative Mismatches \\
 \rowcolor{purple!10}
\textbf{Narrative \& Style} & Perspective Confusions \\
 \rowcolor{purple!10}
 & Tone Inconsistencies \\
 \rowcolor{purple!10}
 & Style Shifts \\
 \bottomrule
\end{tabular}
\caption{Consistency-error taxonomy used by ConStory-Bench, comprising five categories and 19 subtypes.}
\label{tab:error-taxonomy}
\end{table}

\subsection{Automated Error Detection Pipeline}
\label{sec:detection-pipeline}

Building on the task structure and error taxonomy, we introduce \textsc{ConStory-Checker}, an automated LLM-as-judge pipeline for scalable and auditable consistency evaluation. The pipeline consists of four stages~\citep{zheng2023judging}:

\paragraph{Stage 1: Category-Guided Extraction.} 
Narratives are scanned using category-specific prompts across five dimensions (Timeline/Plot, Characterization, World-building, Factual, Narrative Style) to extract contradiction-prone spans.

\paragraph{Stage 2: Contradiction Pairing.} 
Extracted spans are compared pairwise and classified as \emph{Consistent} or \emph{Contradictory}, following CheckEval~\citep{lee2024checkeval} and ProxyQA~\citep{tan2024proxyqa}. This reduces false positives and isolates genuine inconsistencies.

\paragraph{Stage 3: Evidence Chains.} 
For each contradiction, we record: \emph{Reasoning} (why it is a contradiction), \emph{Evidence} (quoted text with positions), and \emph{Conclusion} (error type)~\citep{pereira2024check}.

\paragraph{Stage 4: JSON Reports.} 
Standardized JSON outputs capture quotations, positions, pairings, error categories, and explanations, with all judgments anchored to precise character-level offsets.

We adopt \texttt{o4-mini} as the evaluation model to balance accuracy and efficiency; recent studies confirm strong LLM performance on structured judgment tasks~\citep{chen2024humans}. Complete implementation details are provided in Appendix~\ref{sec:storyverify-details}. This four-stage pipeline forms the foundation for the experiments in Section~\ref{sec:evaluation}.

%% file: sec/4_change_evalution.tex

\section{Evaluation}
\label{sec:evaluation}

We evaluate narrative consistency across four types of systems—proprietary models, open-source models, capability-enhanced models, and agentic writing systems—using ConStory-Bench and ConStory-Checker.

\subsection{Experimental Setup}
\label{sec:experimental-setup}

\paragraph{Models and Data.}
We evaluate a comprehensive set of models spanning four categories. 
Proprietary models are from OpenAI~\cite{gpt5}, Google~\cite{comanici2025gemini}, Anthropic~\cite{anthropic2025claude45},  xAI~\cite{xai2025grok4} and others.
Open-source models cover Qwen~\cite{qwen3}, DeepSeek~\cite{deepseekr1}, GLM~\cite{glm}, Kimi~\cite{kimik2}, and others. 
We also include capability-enhanced models fine-tuned for long-form story generation~\cite{wu2025longwriter, pham2024suri, bai2024longalign} and agent-enhanced systems~\cite{wu2025superwriter, wang2025generating} that employ multi-step generation pipelines. 
Each model generates outputs for all 2{,}000 prompts across the four task scenarios (Figure~\ref{fig:task-prompts}) under comparable settings.

\subsection{Results and Analysis}
\label{sec:results-analysis}

\subsubsection{\textcolor{mplBlue}{RQ1}}
\label{sec:rq1-benchmarks}

\noindent\textbf{\textit{To what extent do current LLMs maintain narrative coherence in ultra-long text generation, and do different models exhibit similar distributions of consistency error types?}}

Benchmarking long-form narrative consistency requires metrics that capture both absolute error rates and relative performance across diverse prompts, yet naive error counting fails to account for length variation and prompt difficulty.

\paragraph{Method.}
We employ two complementary metrics to address these challenges, building on established methodologies for ranking-based evaluation~\cite{liu2023gevalnlgevaluationusing, zheng2023judging}.
Simply counting errors per story unfairly penalizes models that generate longer outputs---a 10K-word story intuitively would have more opportunities for errors than a 2K-word one. To remove this length bias, we introduce \textbf{Consistency Error Density (CED)}, which normalizes errors by output length, measuring errors per ten thousand words for model $m$ on story $i$:
\begin{equation}
\text{CED}_{m,i} = \frac{e_{m,i}}{w_{m,i} / 10000},
\label{eq:ced-rq1}
\end{equation}
where $e_{m,i}$ denotes error count and $w_{m,i}$ word count. Model-level scores average over all stories: $\overline{\text{CED}}_m = \frac{1}{N}\sum_{i=1}^{N}\text{CED}_{m,i}$ (lower is better).
However, CED still does not account for varying prompt difficulty: some prompts inherently elicit more errors across all models. To enable fair cross-model comparison that controls for instance-level difficulty, we introduce \textbf{Group Relative Rank (GRR)}, which ranks models within each prompt group. For each story $i$ with $M_i$ candidate outputs, we define a length-aware quality score
\begin{equation}
Q_{m,i} = \frac{w_{m,i}}{1 + e_{m,i}},
\label{eq:quality-rq1}
\end{equation}
rank all models by $Q_{m,i}$ within the same story $i$, and compute GRR:
\begin{equation}
\text{GRR}_m = \frac{1}{N_m}\sum_{i\in I_m} \mathrm{rank}_i(Q_{m,i}).
\label{eq:grr-rq1}
\end{equation}
Detailed computation examples illustrating these metrics are provided in Appendix~\ref{sec:example-calculation}.

\paragraph{Results \& Answer.}
Table~\ref{tab:comprehensive-performance} shows substantial performance variation across the evaluated models.
\textsc{GPT-5-Reasoning} achieves the lowest CED (0.113) and best GRR (2.80), followed by \textsc{Gemini-2.5-Pro} (CED: 0.305) and \textsc{Claude-Sonnet-4.5} (CED: 0.520, GRR: 4.54).
Among open-source models, \textsc{GLM-4.6} and \textsc{Qwen3-32B} exhibit competitive performance (CED: 0.528--0.537), approaching proprietary-level consistency; moreover, capability-enhanced \textsc{LongWriter-Zero} (CED: 0.669) and agent-enhanced \textsc{SuperWriter} (CED: 0.674) achieve comparable results despite different generation strategies.
These benchmarks show that \textbf{\emph{most models still struggle with long-form narrative consistency and make a considerable number of errors, while GPT-5-Reasoning currently delivers the strongest performance among all evaluated systems.}}
Practically, error analysis reveals \textit{Factual \& Detail Consistency} and \textit{Timeline \& Plot Logic} as dominant failure modes, indicating entity tracking and temporal reasoning remain primary challenges.
Beyond model-level comparisons, task type also affects consistency: \textit{Generation} tasks consistently yield higher CED than \textit{Continuation}, \textit{Expansion}, and \textit{Completion} tasks across most models (Table~\ref{tab:task-type-ced}), suggesting that open-ended creation without prior context poses the greatest consistency challenge.
A comprehensive performance ranking is provided in Appendix~\ref{sec:leaderboard}.

\begin{table*}[t!]
\centering 
\small
\caption{Comprehensive performance on ConStory-Bench. \textbf{CED}: Consistency Error Density (errors per 10K words; lower is better). Category columns show CED breakdown: \textbf{Char.} (Characterization), \textbf{Fact.} (Factual \& Detail Consistency), \textbf{Narr.} (Narrative \& Style), \textbf{Time.} (Timeline \& Plot Logic), \textbf{World} (World-building \& Setting). \textbf{GRR}: Group Relative Rank (lower is better). \textbf{Words}: average output length (words). \textbf{Errors}: average error count per story. \textbf{Total}: number of completed stories. \colorbox[rgb]{.741,.843,.933}{\textbf{Blue}} indicates the best model in each column, \colorbox[rgb]{.886,.937,.855}{Green} indicates the second best, and \colorbox[rgb]{1,.949,.8}{Yellow} indicates the third best. Models with Total below 2,000 indicate prompts refused due to safety filtering.}
\label{tab:comprehensive-performance}
\setlength{\tabcolsep}{2.5pt}
\renewcommand{\arraystretch}{1.2}
\resizebox{\textwidth}{!}{
\begin{tabular}{l|c ccccc|c|>{\centering\arraybackslash}p{1.2cm}>{\centering\arraybackslash}p{1.2cm}>{\centering\arraybackslash}p{1.2cm}}
\toprule 
\multirow{2}{*}{\textbf{Model}} & \multicolumn{6}{c|}{\textbf{CED (errors per 10K words) $\downarrow$}} & \multirow{2}{*}{\textbf{GRR $\downarrow$}} & \multirow{2}{*}{\textbf{Words}} & \multirow{2}{*}{\textbf{Errors}} & \multirow{2}{*}{\textbf{Total}} \\
\cmidrule(lr){2-7}
& \textbf{Overall} & \textbf{Char.} & \textbf{Fact.} & \textbf{Narr.} & \textbf{Time.} & \textbf{World} & & & & \\ 
\midrule 
\multicolumn{11}{c}{\cellcolor[gray]{0.9}\textit{Proprietary Models}} \\ 
\midrule 
GPT-5-Reasoning & \cellcolor[rgb]{.741,.843,.933}\textbf{0.113} & \cellcolor[rgb]{.741,.843,.933}\textbf{0.005} & \cellcolor[rgb]{.741,.843,.933}\textbf{0.061} & \cellcolor[rgb]{.886,.937,.855}0.003 & \cellcolor[rgb]{.741,.843,.933}\textbf{0.024} & \cellcolor[rgb]{.741,.843,.933}\textbf{0.003} & \cellcolor[rgb]{.741,.843,.933}\textbf{3.05} & 9050 & 0.09 & 1990 \\
Gemini-2.5-Pro & \cellcolor[rgb]{.886,.937,.855}0.305 & \cellcolor[rgb]{.886,.937,.855}0.009 & 0.132 & 0.015 & \cellcolor[rgb]{1,.949,.8}0.108 & \cellcolor[rgb]{.886,.937,.855}0.029 & 7.79 & 5584 & 0.16 & 1996 \\
Claude-Sonnet-4.5 & \cellcolor[rgb]{1,.949,.8}0.520 & 0.017 & 0.224 & \cellcolor[rgb]{1,.949,.8}0.004 & 0.128 & 0.043 & \cellcolor[rgb]{.886,.937,.855}4.9 & 8929 & 0.37 & 1998 \\
Grok-4 & 0.670 & 0.033 & 0.307 & 0.065 & 0.222 & 0.076 & 13.38 & 2765 & 0.19 & 2000 \\
GPT-4o-1120 & 0.711 & 0.036 & 0.163 & 0.018 & 0.440 & 0.104 & 17.59 & 1241 & 0.09 & 1774 \\
Doubao-1.6-Thinking-2507 & 1.217 & 0.070 & 0.407 & 0.035 & 0.355 & 0.160 & 11.9 & 3713 & 0.41 & 2000 \\
Mistral-Medium-3.1 & 1.355 & 0.067 & 0.435 & 0.010 & 0.474 & 0.155 & 14.67 & 2447 & 0.28 & 2000 \\
\midrule 
\multicolumn{11}{c}{\cellcolor[gray]{0.9}\textit{Open-source Models}} \\
\midrule 
GLM-4.6 & 0.528 & 0.015 & 0.184 & 0.007 & \cellcolor[rgb]{.886,.937,.855}0.102 & 0.051 & 8.45 & 4949 & 0.18 & 2000 \\
Qwen3-32B & 0.537 & \cellcolor[rgb]{1,.949,.8}0.009 & \cellcolor[rgb]{1,.949,.8}0.120 & 0.068 & 0.191 & 0.047 & 6.39 & 6237 & 0.27 & 2000 \\
Ring-1T & 0.539 & 0.012 & 0.249 & 0.015 & 0.111 & 0.048 & 8.08 & 5264 & 0.23 & 1999 \\
DeepSeek-V3.2-Exp & 0.541 & 0.011 & 0.201 & 0.012 & 0.129 & 0.044 & 10.89 & 3724 & 0.15 & 2000 \\
Qwen3-235B-A22B-Thinking & 0.559 & 0.013 & 0.269 & 0.010 & 0.136 & 0.069 & 7.89 & 5424 & 0.27 & 2000 \\
Step3 & 0.845 & 0.017 & 0.330 & 0.116 & 0.189 & 0.061 & 11.45 & 3793 & 0.27 & 1916 \\
Kimi-K2-2509 & 1.300 & 0.016 & 0.630 & 0.007 & 0.311 & 0.099 & 13.32 & 3227 & 0.34 & 1792 \\
Nvidia-llama-3.1-Ultra & 1.833 & 0.045 & 0.376 & 0.045 & 0.793 & 0.151 & 17.82 & 1224 & 0.17 & 1998 \\
MiniMax-M1-80k & 3.447 & 0.133 & 1.079 & 0.004 & 1.050 & 0.376 & 18.07 & 1442 & 0.38 & 1716 \\
\midrule 
\multicolumn{11}{c}{\cellcolor[gray]{0.9}\textit{Capability-enhanced LLMs}} \\ 
\midrule 
LongWriter-Zero~{\scriptsize\cite{wu2025longwriter}} & 0.669 & 0.027 & \cellcolor[rgb]{.886,.937,.855}0.097 & 0.054 & 0.178 & 0.039 & \cellcolor[rgb]{1,.949,.8}5.45 & 13393 & 0.53 & 1857 \\
Suri-i-ORPO~{\scriptsize\cite{pham2024suri}} & 2.445 & 0.129 & 0.225 & 0.236 & 0.689 & 0.122 & 12.76 & 4279 & 0.60 & 2000 \\
LongAlign-13B-64k~{\scriptsize\cite{bai2024longalign}} & 3.664 & 0.099 & 1.720 & \cellcolor[rgb]{.741,.843,.933}\textbf{0.002} & 0.751 & 0.123 & 18.88 & 1624 & 0.20 & 2000 \\
\midrule 
\multicolumn{11}{c}{\cellcolor[gray]{0.9}\textit{Agent-enhanced Systems}} \\ 
\midrule 
SuperWriter~{\scriptsize\cite{wu2025superwriter}} & 0.674 & 0.025 & 0.255 & 0.070 & 0.245 & \cellcolor[rgb]{1,.949,.8}0.030 & 7.97 & 6036 & 0.38 & 2000 \\
DOME~{\scriptsize\cite{wang2025generating}} & 1.033 & 0.037 & 0.591 & 0.018 & 0.288 & 0.068 & 6.94 & 8399 & 0.84 & 1969 \\
\bottomrule 
\end{tabular}
} 
\end{table*}


\subsubsection{\textcolor{mplOrange}{RQ2}}
\label{sec:rq2-length-dynamics}

\noindent\textbf{\textit{How do consistency errors scale as a function of output length across different LLM architectures?}}

To assess long-form narrative generation, we need to understand how consistency behaves as the generated text grows.
In practice, models that prefer shorter outputs may appear more consistent but leave storylines unfinished, whereas models that write longer texts may complete narratives yet accumulate more contradictions.
To study these patterns, we analyze output length distributions across the evaluated models and examine how error counts scale with increasing narrative length.

\begin{figure}[tb]
\centering
\includegraphics[width=\columnwidth]{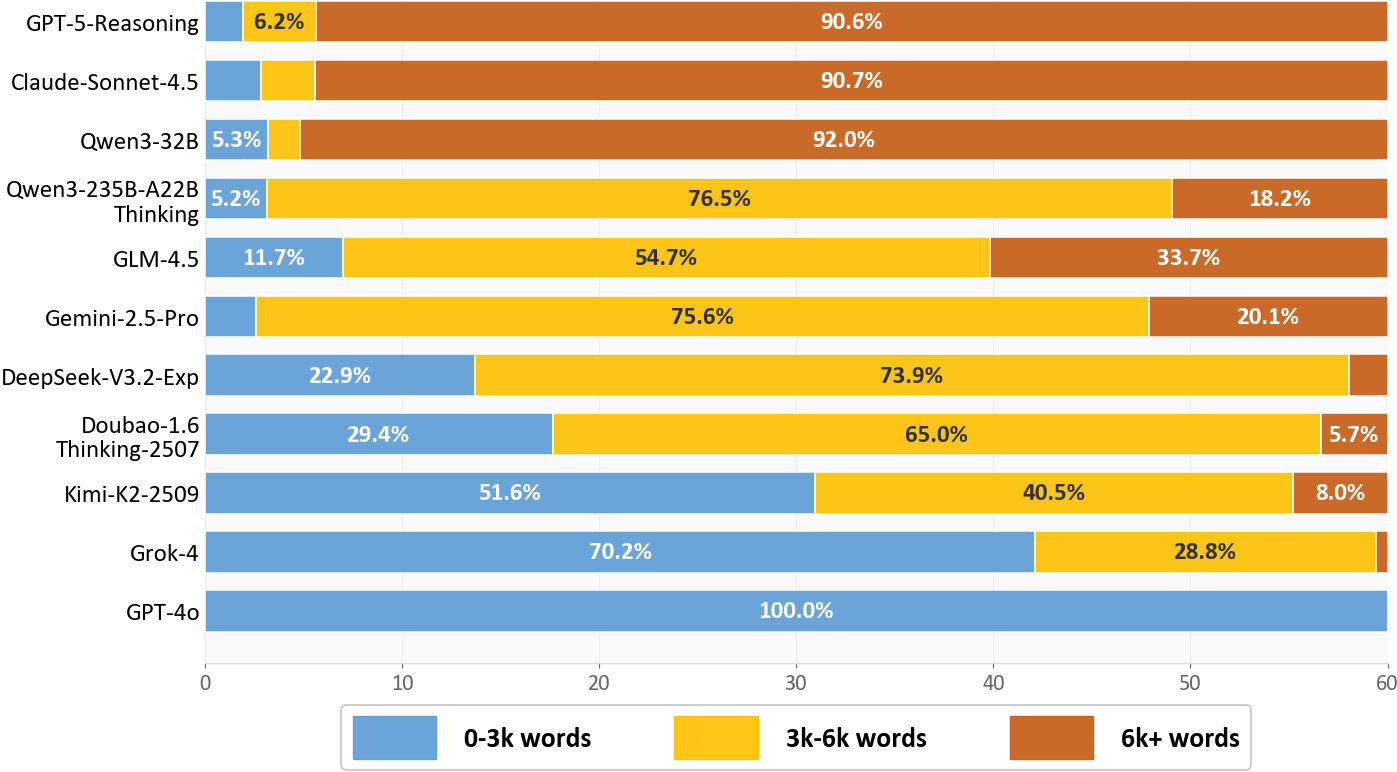}
\caption{Output length distribution across representative models. Stacked bars show the proportion of 0--3K, 3K--6K, and 6K+ word outputs.}
\label{fig:length-distribution}
\end{figure}

\begin{figure}[tb]
\centering
\includegraphics[width=\columnwidth]{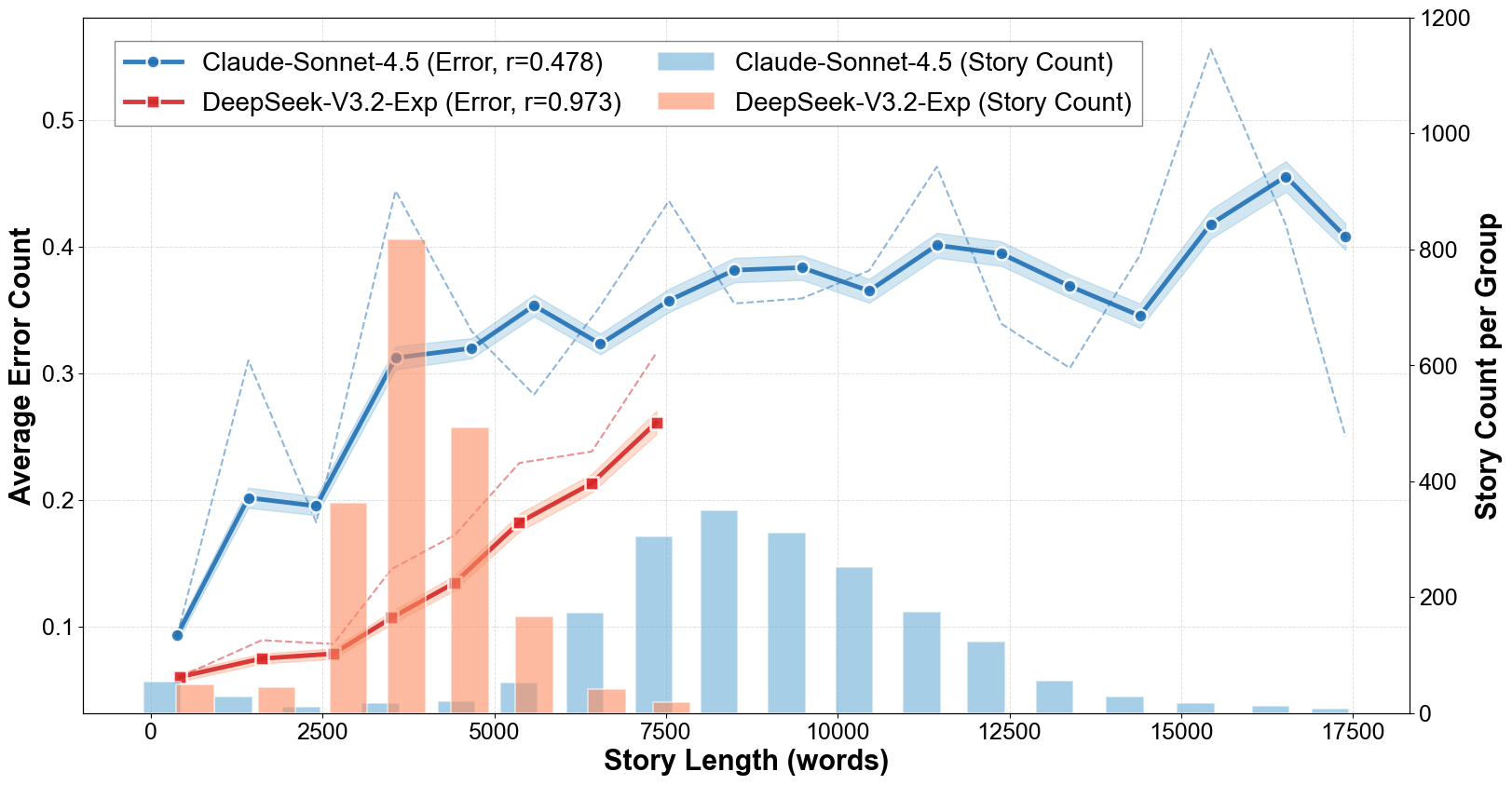}
\caption{Consistency error growth across different story lengths for two models. Lines: Average error count per story at each length bin (cf.\ ``Errors'' in Table~\ref{tab:comprehensive-performance}); Bars: Number of samples in each bin.}
\label{fig:length-error-correlation}
\end{figure}

\paragraph{Results \& Answer.}
Figure~\ref{fig:length-distribution} reveals highly diverse length preferences across evaluated models.
Proprietary systems like \textsc{GPT-5-Reasoning} and \textsc{Claude-Sonnet-4.5} predominantly produce outputs exceeding 6K words (90.6\% and 90.7\% respectively), while \textsc{Grok-4} and \textsc{GPT-4o-1120} predominantly generate shorter outputs, with the majority concentrated in 0--3K words (70.2\% and 100\% respectively).
Open-source models exhibit varied preferences: \textsc{Qwen3-32B} favors longer outputs (92.0\% beyond 3K words), whereas \textsc{DeepSeek-V3.2-Exp} balances across ranges.
As shown in Figure~\ref{fig:length-error-correlation}, error counts increase approximately linearly with output length across models.
\textsc{Claude-Sonnet-4.5} exhibits moderate length-error correlation (r=0.478), while \textsc{DeepSeek-V3.2-Exp} shows stronger dependency (r=0.973).
These patterns demonstrate that \textbf{\emph{errors accumulate linearly with length; however, models differ substantially in their length preferences, leading to diverse length-consistency patterns.}}
Additional model output length statistics are provided in Appendix~\ref{sec:length-statistics}.

\subsubsection{\textcolor{mplPurple}{RQ3}}
\label{sec:rq3-entropy}

\noindent\textbf{\textit{What underlying factors contribute to the emergence of consistency errors, and are there identifiable signals that reliably predict their occurrence?}}

\paragraph{Method.}
We examine whether model \emph{uncertainty} differs between erroneous and correct content. We quantify token-level uncertainty using \textbf{Shannon entropy}~\cite{wang2025beyond, khalid2025ergo, krishnan2024enhancing, lee2025uncertainty}. We select two models, \textsc{Qwen3-4B-Instruct-2507} (4B) and \textsc{Qwen3-30B-A3B-Instruct-2507} (30B), because they are open-source and reproducible, have adequate error samples, and impose manageable computational costs for entropy calculation over long contexts. For each position $t$ in a generated sequence, let $P_t=\{p_1,p_2,\ldots,p_K\}$ denote the next-token distribution over the top-$K$ candidates; then
\begin{equation}
H(P_t)=-\sum_{i=1}^{K} p_i \log_2 p_i,
\end{equation}
where higher entropy signifies a more diffuse, less confident distribution. For a text segment $S$ with $N$ tokens, we report the sentence-level mean
\[
\bar{H}(S)=\frac{1}{N}\sum_{t=1}^{N} H(P_t).
\]
All entropy measurements are based on decoding configurations commonly used in practice: \emph{Temperature} = 0.7, \emph{Top-$k$} = 20, and \emph{Top-$p$} = 0.95. We compute $\bar{H}$ for \emph{error content} and the \emph{whole-text} baseline across representative models.

\paragraph{Results \& Answer.}
Across two representative models, error content exhibits consistently and significantly higher entropy than the whole-text baseline: \textsc{Qwen3-4B-Instruct-2507} shows an entropy increase of 19.24\%, while \textsc{Qwen3-30B-A3B-Instruct-2507} shows an increase of 12.03\% (Table~\ref{tab:entropy-comparison}). Taken together, these results indicate that \textbf{\emph{the model does not err unknowingly; rather, it more often makes incorrect choices when confronted with greater uncertainty.}} Practically, this makes entropy an \emph{actionable} early-warning signal: when local entropy surpasses a stability threshold, the system should trigger verification or self-check routines to curb consistency failures proactively. For complementary token-level uncertainty measures (e.g., probability, perplexity), see Appendix~\ref{sec:token-uncertainty}.


\begin{table}[t!]
\centering
\small
\setlength{\tabcolsep}{4pt}
\renewcommand{\arraystretch}{1.2}
\caption{Entropy comparison between whole text and error-bearing segments. Higher entropy in error content indicates greater unpredictability relative to the whole-text baseline.}
\label{tab:entropy-comparison}
\resizebox{\columnwidth}{!}{
\begin{tabular}{lcc}
\toprule
\textbf{Metric} & \textbf{Qwen3-30B-A3B} & \textbf{Qwen3-4B} \\
 & \textbf{Instruct-2507} & \textbf{Instruct-2507} \\
\midrule
Whole Text Avg Entropy & 1.1438 & 1.0734 \\
Error Content Avg Entropy & 1.2814 & 1.2799 \\
\midrule
Error vs Whole (\% Diff) & \textbf{\textcolor{red!70!black}{+12.03\%}} & \textbf{\textcolor{red!70!black}{+19.24\%}} \\
\bottomrule
\end{tabular}
}
\end{table}


\subsubsection{\textcolor{mplRed}{RQ4}}
\label{sec:rq4-correlations}

\noindent\textbf{\textit{Do different types of consistency errors systematically co-occur, or do they arise independently?}}

Correlation patterns between error categories can reveal important relationships: if two error types often appear together, they may share a common cause; if they rarely co-occur, they likely arise independently.
To quantify these patterns, we compute pairwise Pearson correlation coefficients among the five error categories across all model outputs.

\begin{figure}[t!]
\centering
\includegraphics[width=0.4\textwidth]{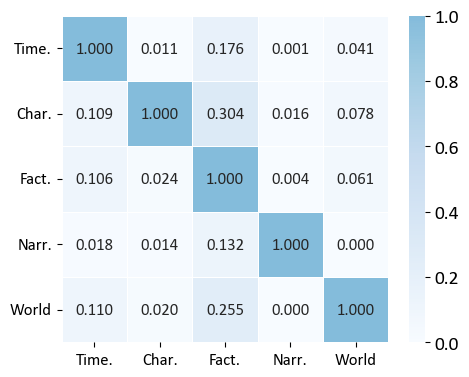}
\caption{Correlation matrix of error categories across all model outputs. Higher values (darker blue) indicate stronger co-occurrence of error types.}
\label{fig:error-correlation}
\end{figure}

\paragraph{Results \& Answer.}
Figure~\ref{fig:error-correlation} shows that \textit{Factual \& Detail Consistency} serves as a central hub, correlating most strongly with \textit{Characterization} (r=0.304), \textit{World-building \& Setting} (r=0.255), and \textit{Timeline \& Plot Logic} (r=0.176).
\textbf{\emph{This heterogeneous correlation structure demonstrates that consistency failures do not arise uniformly; rather, they cluster along specific dependency chains.}}
In contrast, \textit{Narrative \& Style} errors exhibit near-zero correlations with all other categories, indicating that stylistic inconsistencies arise through mechanisms distinct from factual or logical failures.
The strong correlation between \textit{Factual \& Detail Consistency} and other categories suggests these errors tend to co-occur, likely sharing underlying failure mechanisms.
Model-specific correlation patterns are provided in Appendix~\ref{sec:model-error-correlations}.

\subsubsection{\textcolor{mplGreen}{RQ5}}
\label{sec:rq5-positions}

\noindent\textbf{\textit{How are consistency errors distributed across positions within long-form generated narratives?}}

Locating where contradictions appear in the story helps us understand when models start to produce inconsistent content.
The distance between facts and contradictions matters: contradictions shortly after facts suggest local tracking failures, whereas large gaps indicate long-range coherence breakdowns.
To quantify these patterns, we record three normalized positional metrics for each error instance: (1) the position where the original fact is first established (\emph{fact position}), (2) the position where the contradiction appears (\emph{contradiction position}), and (3) the distance between them (\emph{gap}).
The average gap is computed as $\text{Avg Gap} = \frac{1}{n}\sum_{i=1}^{n}|\text{contra}_i - \text{fact}_i|$.
By design, earlier narrative content serves as the ground truth against which later content is evaluated for logical consistency.

\begin{figure*}[t!]
\centering
\includegraphics[width=\textwidth]{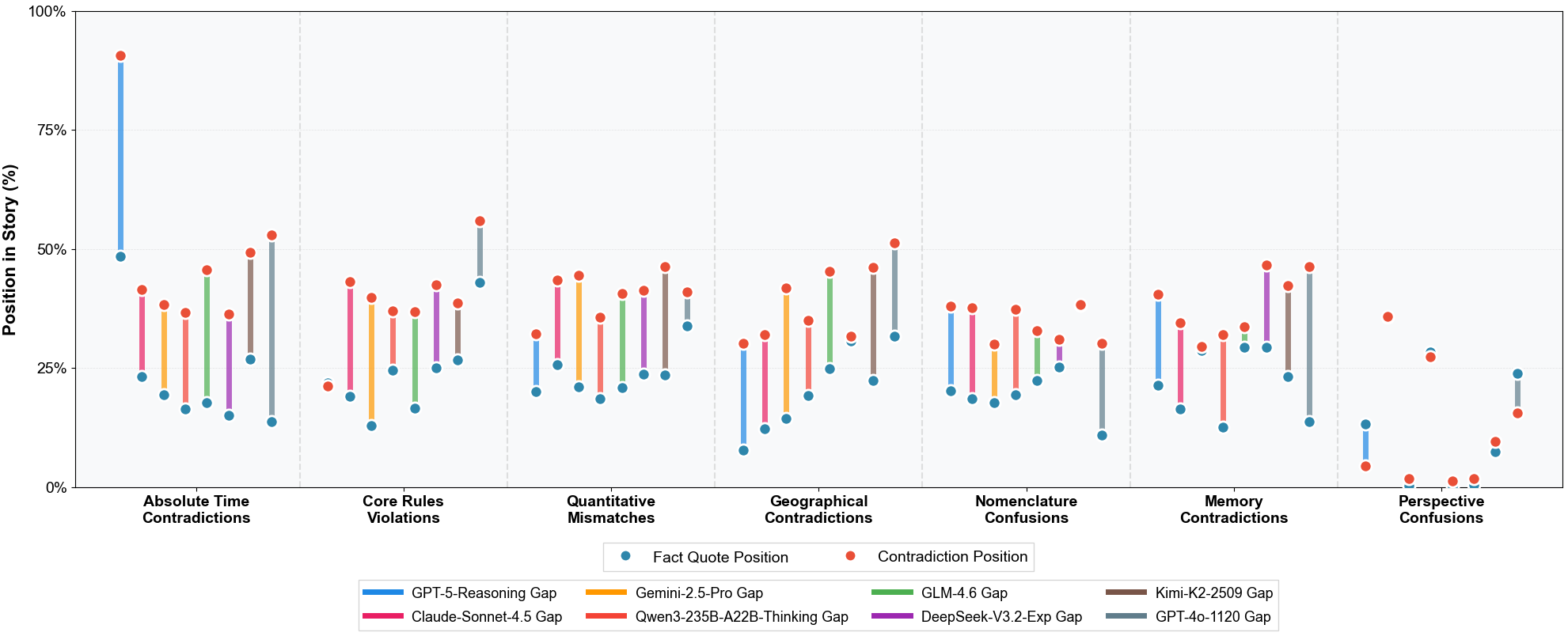}
\caption{Dumbbell plot of error positional distributions. Each row represents an error subtype; blue dots show where facts are first established (fact position), red dots show where contradictions appear, and the connecting line indicates the gap. Columns show four representative models. Values are normalized by story length.}
\label{fig:error-heatmaps}
\end{figure*}

\paragraph{Results \& Answer.}
As shown in Table~\ref{tab:error-position}, geographical contradictions exhibit the largest average positional gap (31.0\%), followed by absolute-time contradictions (29.7\%), while perspective confusions show minimal gaps (4.7\%), suggesting these arise from local rather than long-range context failures.
Figure~\ref{fig:error-heatmaps} visualizes these positional dynamics through dumbbell plots spanning four representative models.
These spatial distributions demonstrate that \textbf{\emph{errors are not uniformly distributed; rather, different error types emerge at characteristic positions along the narrative, with contradiction positions predominantly clustering in the 40--60\% range.}}
Across models, fact positions (blue) concentrate in the early-to-mid narrative (15--30\%), while contradiction positions (red) extend toward later sections.
\textsc{GPT-5-Reasoning} shows the widest gaps for absolute-time contradictions, whereas \textsc{Qwen3-235B-A22B-Thinking} exhibits more compressed gaps overall.
Notably, \textit{perspective confusions} display minimal gaps across all models, suggesting these errors arise from local rather than long-range context failures.
Practically, the systematic gap patterns highlight that temporal and geographical errors require robust long-range memory mechanisms, while stylistic errors may be addressed through local consistency checks.
Extended positional analysis across additional models is provided in Appendix~\ref{sec:extended-positional}.

\begin{table*}[t!]
\centering
\small
\setlength{\tabcolsep}{6pt}
\renewcommand{\arraystretch}{1.25}
\caption{Positional distribution of seven representative error subtypes. Positions are normalized by story length (0--100\%).}
\label{tab:error-position}
\resizebox{\textwidth}{!}{
\begin{tabular}{lccccccc}
\toprule
\textbf{Metric} & \textbf{Absolute Time} & \textbf{Core Rules} & \textbf{Quantitative} & \textbf{Geographical} & \textbf{Nomenclature} & \textbf{Memory} & \textbf{Perspective} \\
 & \textbf{Contradictions} & \textbf{Violations} & \textbf{Mismatches} & \textbf{Contradictions} & \textbf{Confusions} & \textbf{Contradictions} & \textbf{Confusions} \\
\midrule
Avg Fact  & 22.6\% & 23.7\% & 23.4\% & 20.4\% & 21.6\% & 21.8\% & 13.7\% \\
Avg Contradiction & 48.9\% & 39.4\% & 40.6\% & 39.2\% & 34.4\% & 38.2\% & 12.2\% \\
Avg Gap    & 29.7\% & 23.4\% & 23.8\% & 31.0\% & 23.3\% & 25.4\% & 4.7\%  \\
\bottomrule
\end{tabular}
} 
\end{table*}

%% file: sec/2_related.tex
\section{Related Work}

\noindent\textbf{Story Generation.}
Narrative generation tests coherence across plot, characters, and timelines. Planning methods—iterative planning \cite{xie2024creating}, pacing control \cite{wang2023improving}, hierarchical outlines \cite{wang2025generating}, and recurrent mechanisms \cite{zhou2023recurrentgpt}—improve structure; CHIRON \cite{gurung2024chiron} finds character inconsistency. Multi-agent collaboration \cite{huot2024agents} and retrieval-augmented generation \cite{wen2023grove} improve grounding. Length extension methods \cite{bai2024longwriter, pham2024suri, wu2025superwriter, tu2025longwriter} enable longer outputs, but coherence degrades with length \cite{que2024hellobench, wu2024longgenbench}.

\noindent\textbf{Long-Form Generation Benchmarks.}
As context windows expand, long-form evaluation becomes critical. Early work relied on perplexity \cite{beltagy2020longformer, roy2021efficient, press2021train}, which correlates poorly with real use. LongBench \cite{bai2024longbench, bai2025longbench} provides long-context evaluation spanning 8K–2M tokens, while HelloBench \cite{que2024hellobench} and WritingBench \cite{wu2025writingbench} focus on generation quality; models struggle at 16K–32K tokens \cite{wu2024longgenbench}. Classical metrics (ROUGE, BLEU, METEOR) correlate weakly with human judgments \cite{que2024hellobench}, so recent work adds checklist mechanisms \cite{que2024hellobench, lee2024checkeval, pereira2024check}, dynamic criteria \cite{wu2025writingbench}, and proxy-based evaluation \cite{tan2024proxyqa}. Yet many benchmarks rely on fixed templates \cite{paech2023eq, que2024hellobench, bai2025longbench}, limiting fine-grained error detection. For stories, existing evaluations focus on holistic quality \cite{ismayilzada2024evaluating, wang2024weaver, xie2023next} rather than systematic contradiction detection.

%% file: sec/5_conclusion.tex
\section{Conclusion}
\label{sec:conclusion}


We presented \textsc{ConStory-Bench}, a benchmark, and \textsc{ConStory-Checker}, an evaluation pipeline, for assessing narrative consistency in long-form story generation. Our experiments show that current LLMs still produce systematic consistency errors, especially in factual tracking and temporal reasoning; moreover, these errors are not random but cluster in predictable narrative regions. We will provide an interactive portal where the community can discover and submit new consistency errors and checking techniques.


%% file: sec/6_limitations.tex

\section{Limitations}
\label{sec:limitations}

We acknowledge several limitations of this work. First, our benchmark focuses on English fiction following Western narrative conventions. Different cultures have different expectations for storytelling, and we have not evaluated how well ConStory-Bench applies to narratives from other cultural or linguistic backgrounds. Second, we model consistency as a binary judgment—content is either consistent or contradictory. However, some apparent contradictions may serve intentional purposes, such as surprise endings or strategically delayed information; our approach does not distinguish these from true errors. Third, we focus on fiction and storytelling, while long-form consistency is also important in other domains such as technical documentation, academic writing, and screenplays, each with its own conventions.

These limitations suggest several directions for future work: extending the benchmark to multilingual and cross-cultural contexts, developing methods to recognize intentional ambiguity, and adapting the framework to evaluate consistency in other long-form genres.


%% file: sec/appendix.tex
\appendix

\renewcommand{\floatpagefraction}{0.7}
\renewcommand{\topfraction}{0.9}
\renewcommand{\bottomfraction}{0.8}
\renewcommand{\textfraction}{0.1}

\section{Benchmark Construction}
\label{sec:benchmark-construction}

This appendix section details the construction methodology of ConStory-Bench, including task type design rationale, the \textsc{ConStory-Checker} evaluation pipeline, and additional experimental configurations.

\subsection{Task Type Design Rationale}

The four task types are designed to capture different aspects of narrative consistency challenges commonly encountered in long-form generation. Representative prompts for each task type are shown in Figure~\ref{fig:task-prompts}.

\textbf{Generation} involves producing free-form narratives from minimal plot setups, where consistent characters, rules, and causal chains must be instantiated without prior context.

\textbf{Continuation} extends initial story fragments into complete, coherent narratives while preserving established facts, timelines, and character states, ensuring new events remain causally compatible with the given context~\citep{zhou2023recurrentgpt}.

\textbf{Expansion} develops long-form stories from concise yet relatively complete plot outlines by elaborating implicit details and events while maintaining global consistency as narrative complexity increases.

\textbf{Completion} writes full stories with predefined beginnings and endings, filling in the intervening plot to produce coherent and causally well-formed narratives, mirroring collaborative writing workflows.

\begin{figure*}[!t]
\centering
\begin{tcolorbox}[width=0.95\textwidth, title=Example Prompts for Four Task Types, colback=blue!5, colframe=blue!30]
\small  

\textbf{System Prompt (Shared across all task types):}

\textit{``You are a master storyteller and creative writing expert. Your task is to generate high-quality, engaging stories based on the given prompts.}

\textit{Create compelling narratives with: Rich character development; Vivid descriptions and settings; Engaging plot progression; Appropriate tone and style for the prompt; Well-structured storytelling; STRICTLY follow any word count or length requirements specified in the prompt.}

\textit{Generate complete, well-crafted stories that bring the prompts to life with creativity and literary skill. Pay careful attention to any specific length, word count, or format requirements mentioned in the prompt and ensure your story meets those exact specifications.''}

\medskip
\hrule
\medskip

\textbf{[Task Type 1: Generation]}

Hey, can you write me a new story about a dog who's been living it up on table scraps and now has to switch to a strict diet because of chronic pancreatitis? I want it to really dig into themes of sacrifice, empathy, and the bond between pet and owner, with memorable scenes at the vet's office, at mealtime and during a tense overnight crisis. Set it in a cozy small town and show how the owner adjusts to the low-protein routine. Shoot for about 8,000 to 10,000 words.

\medskip
\textbf{[Task Type 2: Continuation]}

Write a story about two online friends who lure a teammate camping a sniper rifle into accidentally team-killing one of them, get him kicked, and then obsessively follow his trail across multiple servers. Start with the tense Urban Terror match on that map with the hot-dog cart, showing their knife-throwing prank and fake apologies before he finally snaps. Explore their motivations, the thrill of the chase across different servers, and how this prank tests their friendship and sense of empathy. The story should be roughly 8,000 to 10,000 words.

\medskip
\textbf{[Task Type 3: Expansion]}

Turn this wild delivery story into an 8,000--10,000 word narrative: a shift manager at a Papa John's in rural Alaska steps in to make a late-night pizza run after drivers call in sick and runs headfirst into a bull moose and a separated mother-calf pair under a motion-sensor floodlight. Flesh out the manager's backstory and fears, paint the Alaskan night with vivid sensory details, and build mounting tension as he navigates the encounter. Include his internal monologue or dialogue with the assistant manager back at the shop, and finish with his emotional reflection on the near-death experience and getting stiffed on the tip.

\medskip
\textbf{[Task Type 4: Completion]}

Write a story about a guy in his early twenties who once dated a girl in his college friend group and then watched her start seeing his best friend soon after they split. Start your story at a lively spring party where he meets a new girl from that same circle of friends, completely unaware of the past drama. End with him standing in the moonlit quad, having made his decision about whether to confess his history with his ex to this new girlfriend. Include scenes of introspection, a tense run-in with his ex, and a heartfelt conversation under the stars, exploring themes of honesty, trust, and forgiveness. The finished story should be roughly 8,000--10,000 words.
\end{tcolorbox}
\caption{Example prompts for the four narrative generation task types in ConStory-Bench.}
\label{fig:task-prompts}
\end{figure*}


\subsection{ConStory-Checker: Detailed Implementation}
\label{sec:storyverify-details}

Extending the conceptual framework presented in Section~\ref{sec:detection-pipeline}, this subsection provides comprehensive implementation details of the \textsc{ConStory-Checker} evaluation pipeline. While Section~\ref{sec:detection-pipeline} introduces the four-stage detection process, a complete specification of prompt structures, category taxonomies, and output schemas is essential for reproducible consistency evaluation at scale.

\paragraph{Category Definitions.}
The \textsc{ConStory-Checker} pipeline evaluates narrative consistency through five complementary error dimensions (Figures~\ref{fig:storyverify-timeline}--\ref{fig:storyverify-narrative}): \textit{Timeline \& Plot Logic}, \textit{Characterization}, \textit{World-building \& Setting}, \textit{Factual \& Detail Consistency}, and \textit{Narrative \& Style}. Each category employs structured extraction guidelines with standardized JSON output schemas specifying \texttt{fact\_quote}, \texttt{location}, \texttt{contradiction\_pair}, \texttt{error\_element}, \texttt{error\_category}, and \texttt{context} fields, enabling systematic cross-document comparison and aggregation.

\paragraph{Validation Methodology.}
To empirically validate \textsc{ConStory-Checker}'s effectiveness, we constructed a diagnostic dataset through systematic error injection into authentic narrative contexts. Using \texttt{Qwen3-235B-A22B-Thinking}, we generated 200 stories with deliberately planted inconsistencies across all five error dimensions (1,000 injected errors total). Two professional web novel writers independently annotated this dataset at \$1.00 per story, completing all 200 stories within two days and establishing human expert baselines for comparison. The annotation protocol required annotators to first study the five-category error taxonomy and subtype definitions, then read each story in full to identify consistency errors. For each error, they recorded: (1) the fact quote, (2) the contradicting quote, (3) the error category and subtype, and (4) a brief explanation of the inconsistency. We evaluated both \textsc{ConStory-Checker} (Direct Detection) and human annotators (Human Detection) using standard classification metrics---\textbf{Precision}, \textbf{Recall}, and \textbf{F1-score}---with the injected errors serving as ground truth.

\begin{figure*}[!t]
\centering
\includegraphics[width=\textwidth]{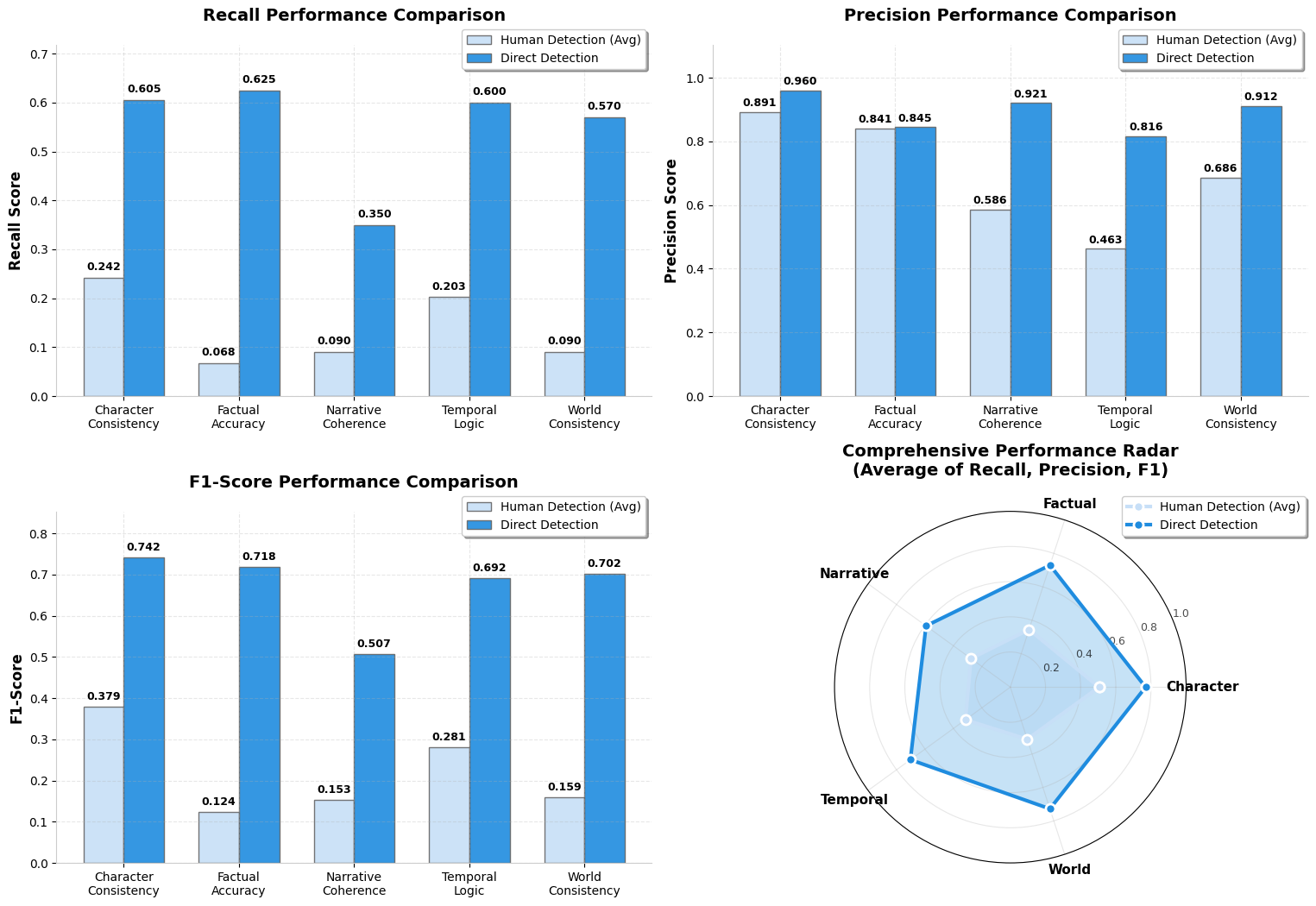}
\caption{Performance comparison between \textsc{ConStory-Checker} and human expert judgment across five consistency error categories. The evaluation covers recall, precision, F1-score, and comprehensive radar visualization, demonstrating that our automated approach achieves human-competitive performance in detecting narrative inconsistencies.}
\label{fig:storyverify-performance}
\end{figure*}

\paragraph{Results.}
Table~\ref{tab:storyverify-performance} and Figure~\ref{fig:storyverify-performance} present performance comparisons across all error categories. The results demonstrate that \textbf{\textsc{ConStory-Checker} (Overall F1=0.678) substantially outperforms human expert judgment (Overall F1=0.281) in detecting narrative inconsistencies}. The automated system achieves high precision (0.884) while maintaining robust recall (0.550), with particularly strong performance in \textbf{Character Consistency} (F1=0.742) and \textbf{Factual Accuracy} (F1=0.718). In contrast, human annotators exhibit substantially lower recall across all dimensions---ranging from 4.5\% to 31.5\%. Notably, \textsc{ConStory-Checker} detects 550 of 1,000 injected errors (55.0\% recall) compared to only 171 detections by human experts (17.1\% recall), representing a \textbf{3.2$\times$ improvement in error discovery rate}. Figure~\ref{fig:storyverify-examples} illustrates representative detection cases, demonstrating \textsc{ConStory-Checker}'s capability to identify subtle contradictions across all five dimensions. These findings validate that automated consistency evaluation provides more comprehensive and reliable detection of narrative consistency errors compared to manual human judgment.

\begin{table*}[!t]
\centering
\small
\caption{Performance comparison between \textsc{ConStory-Checker} (Direct Detection) and human expert judgment (Human Detection) on the diagnostic dataset. GT (Ground Truth) indicates the 200 injected errors for each category. The error categories are: \textbf{Char.} (Characterization), \textbf{Fact.} (Factual \& Detail Consistency), \textbf{Narr.} (Narrative \& Style), \textbf{Time.} (Timeline \& Plot Logic), and \textbf{World} (World-building \& Setting).}
\label{tab:storyverify-performance}
\setlength{\tabcolsep}{6pt}
\renewcommand{\arraystretch}{1.1}
\begin{tabular}{lc|ccccc|ccccc}
\toprule
\multirow{2}{*}{\textbf{Error Category}} & \multirow{2}{*}{\textbf{GT}} & \multicolumn{5}{c|}{\textbf{Direct Detection}} & \multicolumn{5}{c}{\textbf{Experts Detection (Avg)}} \\
\cmidrule(lr){3-7} \cmidrule(lr){8-12}
& & \textbf{Pred} & \textbf{TP} & \textbf{Recall} & \textbf{Prec} & \textbf{F1} & \textbf{Pred} & \textbf{TP} & \textbf{Recall} & \textbf{Prec} & \textbf{F1} \\
\midrule
\textbf{Character Consistency} & 200 & 126 & 121 & 0.605 & 0.960 & 0.742 & 53.5 & 48.5 & 0.242 & 0.891 & 0.379 \\
\textbf{Factual Accuracy} & 200 & 148 & 125 & 0.625 & 0.845 & 0.718 & 16.5 & 13.5 & 0.068 & 0.841 & 0.124 \\
\textbf{Narrative Coherence} & 200 & 76  & 70  & 0.350 & 0.921 & 0.507 & 29.5 & 18.0 & 0.090 & 0.586 & 0.153 \\
\textbf{Temporal Logic} & 200 & 147 & 120 & 0.600 & 0.816 & 0.692 & 84.0 & 40.5 & 0.203 & 0.463 & 0.281 \\
\textbf{World Consistency} & 200 & 125 & 114 & 0.570 & 0.912 & 0.702 & 26.5 & 18.0 & 0.090 & 0.686 & 0.159 \\
\midrule
\textbf{Total} & \textbf{1000} & \textbf{622} & \textbf{550} & \textbf{0.550} & \textbf{0.884} & \textbf{0.678} & \textbf{210.0} & \textbf{138.5} & \textbf{0.139} & \textbf{0.660} & \textbf{0.229} \\
\bottomrule
\end{tabular}
\end{table*}

\begin{figure*}[htbp]
\centering
\begin{tcolorbox}[width=0.95\textwidth, title=ConStory-Checker Evaluation: Timeline \& Plot Logic Category, colback=gray!5]
\scriptsize
\setlength{\parskip}{2pt}
\setlength{\itemsep}{0pt}
\setlength{\parsep}{0pt}
\setlength{\topsep}{0pt}
\setlength{\partopsep}{0pt}
\setlength{\baselineskip}{9.5pt}

\textbf{EXTRACTION CATEGORIES}
\vspace{2pt}

\textbf{Category 1 --- Absolute Time Contradictions}

Extract sentences that contain conflicting calendar information (dates, weekdays, seasons, celestial events).
\vspace{1pt}

\textit{Look for:} (1) Contradictory date statements about the same event; (2) Seasonal descriptions that conflict within a short timeframe; (3) Astronomical references that contradict each other; (4) Weather/climate descriptions inconsistent with stated time periods.
\vspace{3pt}

\textbf{Category 2 --- Duration/Timeline Contradictions}

Extract sentences that give conflicting time measurements for the same events or journeys.
\vspace{1pt}

\textit{Look for:} (1) Different duration statements for identical journeys/events; (2) Travel times that contradict each other; (3) Age progressions that don't match stated time passages; (4) Event sequences with impossible timing.
\vspace{3pt}

\textbf{Category 3 --- Simultaneity Contradictions}

Extract sentences showing characters/objects in multiple locations at the same time.
\vspace{1pt}

\textit{Look for:} (1) Character described as present in different places simultaneously; (2) Objects appearing in multiple locations at once; (3) Dialogue or actions suggesting impossible simultaneous presence; (4) Timeline overlaps that create spatial paradoxes.
\vspace{3pt}

\textbf{Category 4 --- Causeless Effects}

Extract sentences describing major outcomes, skills, or resources that appear without prior setup.
\vspace{1pt}

\textit{Look for:} (1) Characters suddenly possessing unexplained abilities; (2) Important items appearing without introduction; (3) Knowledge or skills emerging without learning/acquisition scenes; (4) Plot developments lacking causal foundation.
\vspace{3pt}

\textbf{Category 5 --- Causal Logic Violations}

Extract sentences where causes and effects contradict established story rules or logic.
\vspace{1pt}

\textit{Look for:} (1) Effects disproportionate to their stated causes; (2) Cause-effect chains that violate established world physics/rules; (3) Character reactions inconsistent with established personalities; (4) World-building contradictions in action-consequence relationships.
\vspace{3pt}

\textbf{Category 6 --- Abandoned Plot Elements}

Extract sentences that introduce significant plot elements which are never resolved or referenced again.
\vspace{1pt}

\textit{Look for:} (1) Important objects/mysteries introduced but never mentioned again; (2) Character goals/motivations that disappear without resolution; (3) Promised revelations or confrontations that never occur; (4) Significant relationships/conflicts that vanish from the narrative.
\vspace{4pt}

\textbf{OUTPUT SCHEMA}
\vspace{1pt}

\textit{Extraction Format:}

For each contradiction found, provide:
\vspace{-1pt}
\begin{itemize}\setlength{\itemsep}{1pt}\setlength{\parskip}{0pt}\setlength{\topsep}{1pt}
\item \textbf{fact\_quote}: The complete contradictory sentence(s) from the original text
\item \textbf{location}: Chapter/paragraph/line reference where found
\item \textbf{contradiction\_pair}: If applicable, the conflicting statement(s) from elsewhere in the text
\item \textbf{contradiction\_location}: Location of the contradicting statement (if contradiction\_pair exists)
\item \textbf{error\_element}: The specific timeline/plot element involved (e.g., "seasonal timing", "journey duration", "character location")
\item \textbf{error\_category}: The broad error type (e.g., "absolute\_time\_error", "duration\_error", "simultaneity\_paradox")
\item \textbf{context}: Brief explanation of why these passages contradict each other
\end{itemize}
\vspace{2pt}

\textit{JSON Output Format:}

\begin{verbatim}
{
  "absolute_time_contradictions": [{
    "fact_quote": "It was a scorching July afternoon...",
    "location": "Chapter 3, paragraph 2",
    "contradiction_pair": "Snow began falling heavily...",
    "contradiction_location": "Chapter 3, paragraph 5",
    "error_element": "seasonal weather timing",
    "error_category": "absolute_time_error",
    "context": "Same day described as both midsummer..."
  }],
  "duration_contradictions": [{...}],
  "simultaneity_contradictions": [{...}],
  "causeless_effects": [{...}],
  "causal_logic_violations": [{...}],
  "abandoned_plot_elements": [{...}]
}
\end{verbatim}

\end{tcolorbox}
\caption{Complete judge prompt for Timeline \& Plot Logic category in \textsc{ConStory-Checker} evaluation protocol.}
\label{fig:storyverify-timeline}
\end{figure*}

\begin{figure*}[htbp]
\centering
\begin{tcolorbox}[width=0.95\textwidth, title=ConStory-Checker Evaluation: Characterization Category, colback=gray!5]
\scriptsize
\setlength{\parskip}{2pt}
\setlength{\itemsep}{0pt}
\setlength{\parsep}{0pt}
\setlength{\topsep}{0pt}
\setlength{\partopsep}{0pt}
\setlength{\baselineskip}{9.5pt}

\textbf{EXTRACTION CATEGORIES}
\vspace{2pt}

\textbf{Category 1 --- Memory Contradictions}

Extract sentences showing characters forgetting crucial personal details or remembering non-existent events.
\vspace{1pt}

\textit{Look for:} (1) Characters forgetting previously established relationships, experiences, or commitments; (2) Characters recalling events, people, or places never mentioned before in the narrative; (3) Inconsistent memories about personal history, family, or past events; (4) Characters acting as if they've never met someone they clearly interacted with before.
\vspace{3pt}

\textbf{Category 2 --- Knowledge Contradictions}

Extract sentences showing characters displaying knowledge, vocabulary, or skills beyond their established background.
\vspace{1pt}

\textit{Look for:} (1) Characters using terminology, concepts, or language inconsistent with their education/background; (2) Medieval characters using modern concepts or vocabulary; (3) Uneducated characters demonstrating advanced technical knowledge; (4) Characters knowing information they couldn't have accessed given their background; (5) Cultural/temporal knowledge mismatches (modern slang in historical settings, etc.).
\vspace{3pt}

\textbf{Category 3 --- Skill/Power Fluctuations}

Extract sentences showing dramatic, unexplained changes in character abilities or competence levels.
\vspace{1pt}

\textit{Look for:} (1) Characters suddenly losing or gaining abilities without explanation; (2) Dramatic power level changes between scenes; (3) Competence levels varying wildly without narrative justification; (4) Characters performing feats far beyond or below their established capabilities; (5) Unexplained shifts in physical, mental, or magical abilities.
\vspace{3pt}

\textbf{Category 4 --- Forgotten Abilities}

Extract sentences showing characters failing to use established abilities when they would logically resolve conflicts.
\vspace{1pt}

\textit{Look for:} (1) Characters with established powers/skills not using them in relevant situations; (2) Abilities mentioned early in the story but ignored in later conflicts; (3) Characters struggling with problems their known abilities could easily solve; (4) Magical/special powers that disappear when they would be most useful; (5) Skills or knowledge that characters "forget" they possess.
\vspace{4pt}

\textbf{OUTPUT SCHEMA}
\vspace{1pt}

\textit{Extraction Format:}

For each contradiction found, provide:
\vspace{-1pt}
\begin{itemize}\setlength{\itemsep}{1pt}\setlength{\parskip}{0pt}\setlength{\topsep}{1pt}
\item \textbf{fact\_quote}: The complete contradictory sentence(s) from the original text
\item \textbf{location}: Chapter/paragraph/line reference where found
\item \textbf{contradiction\_pair}: If applicable, the conflicting statement(s) from elsewhere in the text
\item \textbf{contradiction\_location}: Location of the contradicting statement (if contradiction\_pair exists)
\item \textbf{error\_element}: The specific character element involved (e.g., "character memory", "knowledge background", "ability level", "character name")
\item \textbf{error\_category}: The broad error type (e.g., "memory\_contradiction", "knowledge\_contradiction", "skill\_fluctuation")
\item \textbf{context}: Brief explanation of why these passages show character inconsistency
\end{itemize}
\vspace{2pt}

\textit{JSON Output Format:}

\begin{verbatim}
{
  "memory_contradictions": [{
    "fact_quote": "I've never seen this woman...",
    "location": "Chapter 5, paragraph 3",
    "contradiction_pair": "Sarah and I spent...",
    "contradiction_location": "Chapter 2, paragraph 8",
    "error_element": "character relationship memory",
    "error_category": "memory_contradiction",
    "context": "Character claims not to know someone..."
  }],
  "knowledge_contradictions": [{...}],
  "skill_power_fluctuations": [{...}],
  "forgotten_abilities": [{...}]
}
\end{verbatim}

\end{tcolorbox}
\caption{Complete judge prompt for Characterization category in \textsc{ConStory-Checker} evaluation protocol.}
\label{fig:storyverify-case}
\end{figure*}

\begin{figure*}[htbp]
\centering
\begin{tcolorbox}[width=0.95\textwidth, title=ConStory-Checker Evaluation: World-building \& Setting Category, colback=gray!5]
\scriptsize
\setlength{\parskip}{2pt}
\setlength{\itemsep}{0pt}
\setlength{\parsep}{0pt}
\setlength{\topsep}{0pt}
\setlength{\partopsep}{0pt}
\setlength{\baselineskip}{9.5pt}

\textbf{EXTRACTION CATEGORIES}
\vspace{2pt}

\textbf{Category 1 --- Core Rules Violations}

Extract sentences showing actions that directly violate the author's explicitly stated fundamental world laws or physics.
\vspace{1pt}

\textit{Look for:} (1) Characters performing actions impossible under established world physics; (2) Magic systems being violated or ignored by their own rules; (3) Technology working beyond its established limitations; (4) Natural laws being broken without explanation or justification; (5) World mechanics functioning inconsistently with their established parameters; (6) Established limitations being ignored when convenient for plot.
\vspace{3pt}

\textbf{Category 2 --- Social Norms Violations}

Extract sentences showing character behaviors that contradict established social systems without appropriate consequences.
\vspace{1pt}

\textit{Look for:} (1) Characters violating clearly established laws without facing consequences; (2) Social hierarchies being ignored without realistic reactions; (3) Cultural taboos being broken with no societal response; (4) Religious or ideological systems being contradicted without backlash; (5) Economic systems working inconsistently with established rules; (6) Characters behaving outside their established social roles without explanation.
\vspace{3pt}

\textbf{Category 3 --- Geographical Contradictions}

Extract sentences showing geographical entities changing properties or relative positions inconsistently.
\vspace{1pt}

\textit{Look for:} (1) Cities, mountains, rivers changing location between references; (2) Distances between locations varying dramatically without explanation; (3) Geographical features changing size, appearance, or characteristics; (4) Climate or terrain descriptions contradicting previous descriptions; (5) Travel times inconsistent with established distances; (6) Maps or directional references contradicting each other.
\vspace{4pt}

\textbf{OUTPUT SCHEMA}
\vspace{1pt}

\textit{Extraction Format:}

For each contradiction found, provide:
\vspace{-1pt}
\begin{itemize}\setlength{\itemsep}{1pt}\setlength{\parskip}{0pt}\setlength{\topsep}{1pt}
\item \textbf{fact\_quote}: The complete contradictory sentence(s) from the original text
\item \textbf{location}: Chapter/paragraph/line reference where found
\item \textbf{contradiction\_pair}: If applicable, the conflicting statement(s) from elsewhere in the text
\item \textbf{contradiction\_location}: Location of the contradicting statement (if contradiction\_pair exists)
\item \textbf{error\_element}: The specific world-building element involved (e.g., "magic system rules", "social hierarchy", "geographical location")
\item \textbf{error\_category}: The broad error type (e.g., "core\_rules\_violation", "social\_norms\_violation", "geographical\_contradiction")
\item \textbf{context}: Brief explanation of why these passages show world-building inconsistency
\end{itemize}
\vspace{2pt}

\textit{JSON Output Format:}

\begin{verbatim}
{
  "core_rules_violations": [{
    "fact_quote": "Sarah effortlessly cast three high-level...",
    "location": "Chapter 8, paragraph 4",
    "contradiction_pair": "Even a single major spell drains...",
    "contradiction_location": "Chapter 2, paragraph 15",
    "error_element": "magic system energy limitations",
    "error_category": "core_rules_violation",
    "context": "Character violates established magic system..."
  }],
  "social_norms_violations": [{
    "fact_quote": "The peasant boy openly insulted the king...",
    "location": "Chapter 5, paragraph 12",
    "contradiction_pair": "In our kingdom, any insult to the crown...",
    "contradiction_location": "Chapter 1, paragraph 8",
    "error_element": "royal authority and punishment system",
    "error_category": "social_norms_violation",
    "context": "Character violates established social hierarchy..."
  }],
  "geographical_contradictions": [{...}]
}
\end{verbatim}

\end{tcolorbox}
\caption{Complete judge prompt for World-building \& Setting category in \textsc{ConStory-Checker} evaluation protocol.}
\label{fig:storyverify-world}
\end{figure*}

\begin{figure*}[htbp]
\centering
\begin{tcolorbox}[width=0.95\textwidth, title=ConStory-Checker Evaluation: Factual \& Detail Consistency Category, colback=gray!5]
\scriptsize
\setlength{\parskip}{2pt}
\setlength{\itemsep}{0pt}
\setlength{\parsep}{0pt}
\setlength{\topsep}{0pt}
\setlength{\partopsep}{0pt}
\setlength{\baselineskip}{9.5pt}

\textbf{EXTRACTION CATEGORIES}
\vspace{2pt}

\textbf{Category 1 --- Appearance Mismatches}

Extract sentences showing physical descriptions of the same character or object that change inconsistently.
\vspace{1pt}

\textit{Look for:} (1) Character physical features (hair color, eye color, height, build) described differently; (2) Distinctive marks, scars, or identifying features that appear/disappear; (3) Object appearances (color, size, shape, material) changing between references; (4) Location descriptions varying dramatically for the same place; (5) Clothing or equipment descriptions that contradict previous descriptions; (6) Age-related appearance changes that don't match stated time progression.
\vspace{3pt}

\textbf{Category 2 --- Nomenclature Confusions}

Extract sentences showing names of entities (people, places, objects) used incorrectly or inconsistently.
\vspace{1pt}

\textit{Look for:} (1) Character names spelled differently or changed entirely; (2) Characters referred to by wrong names or titles; (3) Place names varying in spelling or completely changing; (4) Object names, brands, or proper nouns used inconsistently; (5) Family relationships with conflicting names (same person called different names); (6) Titles, ranks, or positions assigned inconsistently to same characters.
\vspace{3pt}

\textbf{Category 3 --- Quantitative Mismatches}

Extract sentences showing numerical information that is mathematically inconsistent.
\vspace{1pt}

\textit{Look for:} (1) Age contradictions (character ages not matching time progression); (2) Date inconsistencies (events happening in impossible chronological order); (3) Quantity changes (same groups, amounts, or measurements given different numbers); (4) Distance and measurement contradictions; (5) Currency, pricing, or economic figure inconsistencies; (6) Population, army sizes, or statistical data contradictions.
\vspace{4pt}

\textbf{OUTPUT SCHEMA}
\vspace{1pt}

\textit{Extraction Format:}

For each contradiction found, provide:
\vspace{-1pt}
\begin{itemize}\setlength{\itemsep}{1pt}\setlength{\parskip}{0pt}\setlength{\topsep}{1pt}
\item \textbf{fact\_quote}: The complete contradictory sentence(s) from the original text
\item \textbf{location}: Chapter/paragraph/line reference where found
\item \textbf{contradiction\_pair}: If applicable, the conflicting statement(s) from elsewhere in the text
\item \textbf{contradiction\_location}: Location of the contradicting statement (if contradiction\_pair exists)
\item \textbf{error\_element}: The specific detail element involved (e.g., "character eye color", "character name", "army size")
\item \textbf{error\_category}: The broad error type (e.g., "appearance\_mismatch", "nomenclature\_confusion", "quantitative\_mismatch")
\item \textbf{context}: Brief explanation of why these passages show factual inconsistency
\end{itemize}
\vspace{2pt}

\textit{JSON Output Format:}

\begin{verbatim}
{
  "appearance_mismatches": [{
    "fact_quote": "Elena's striking emerald green eyes...",
    "location": "Chapter 4, paragraph 2",
    "contradiction_pair": "Her deep brown eyes reflected...",
    "contradiction_location": "Chapter 8, paragraph 15",
    "error_element": "character eye color",
    "error_category": "appearance_mismatch",
    "context": "Same character described with different eye colors..."
  }],
  "nomenclature_confusions": [{
    "fact_quote": "Captain Richardson led his troops...",
    "location": "Chapter 6, paragraph 8",
    "contradiction_pair": "Captain Robinson shouted orders...",
    "contradiction_location": "Chapter 6, paragraph 12",
    "error_element": "character surname",
    "error_category": "nomenclature_confusion",
    "context": "Same military leader referred to by two different..."
  }],
  "quantitative_mismatches": [{...}]
}
\end{verbatim}

\end{tcolorbox}
\caption{Complete judge prompt for Factual \& Detail Consistency category in \textsc{ConStory-Checker} evaluation protocol.}
\label{fig:storyverify-factual}
\end{figure*}

\begin{figure*}[htbp]
\centering
\begin{tcolorbox}[width=0.95\textwidth, title=ConStory-Checker Evaluation: Narrative \& Style Category, colback=gray!5]
\scriptsize
\setlength{\parskip}{2pt}
\setlength{\itemsep}{0pt}
\setlength{\parsep}{0pt}
\setlength{\topsep}{0pt}
\setlength{\partopsep}{0pt}
\setlength{\baselineskip}{9.5pt}

\textbf{EXTRACTION CATEGORIES}
\vspace{2pt}

\textbf{Category 1 --- Perspective Confusions}

Extract sentences showing inappropriate narrative perspective shifts between first-person, third-person, or other viewpoints.
\vspace{1pt}

\textit{Look for:} (1) Sudden shifts from first-person (``I'') to third-person (``he/she'') within scenes; (2) Inconsistent point-of-view within the same paragraph or chapter; (3) Unclear or unjustified perspective changes mid-narrative; (4) Multiple conflicting viewpoints presented simultaneously without clear structure; (5) Omniscient narrator suddenly becoming limited or vice versa; (6) Character thoughts accessible then inaccessible without explanation.
\vspace{3pt}

\textbf{Category 2 --- Tone Inconsistencies}

Extract sentences showing inappropriate tonal shifts without narrative justification.
\vspace{1pt}

\textit{Look for:} (1) Serious dramatic moments interrupted by inappropriate humor; (2) Formal language mixed with colloquial expressions inappropriately; (3) Genre tone violations (comedy elements in horror scenes, etc.); (4) Character voice inconsistencies (formal character suddenly using slang); (5) Mood whiplash between adjacent sentences or paragraphs; (6) Register mismatches (academic language in casual dialogue).
\vspace{3pt}

\textbf{Category 3 --- Style Shifts}

Extract sentences showing dramatic writing style changes without plot-driven necessity.
\vspace{1pt}

\textit{Look for:} (1) Vocabulary sophistication level changing dramatically between sections; (2) Sentence structure patterns shifting unexpectedly (complex to simple or vice versa); (3) Descriptive approach changing without narrative reason; (4) Writing quality or competence appearing to vary significantly; (5) Literary devices used inconsistently or inappropriately; (6) Authorial voice seeming to change between chapters or sections.
\vspace{4pt}

\textbf{OUTPUT SCHEMA}
\vspace{1pt}

\textit{Extraction Format:}

For each contradiction found, provide:
\vspace{-1pt}
\begin{itemize}\setlength{\itemsep}{1pt}\setlength{\parskip}{0pt}\setlength{\topsep}{1pt}
\item \textbf{fact\_quote}: The complete contradictory sentence(s) from the original text
\item \textbf{location}: Chapter/paragraph/line reference where found
\item \textbf{contradiction\_pair}: If applicable, the conflicting statement(s) from elsewhere in the text
\item \textbf{contradiction\_location}: Location of the contradicting statement (if contradiction\_pair exists)
\item \textbf{error\_element}: The specific style element involved (e.g., "point of view shift", "tonal inconsistency", "vocabulary complexity")
\item \textbf{error\_category}: The broad error type (e.g., "perspective\_confusion", "tone\_inconsistency", "style\_shift")
\item \textbf{context}: Brief explanation of why these passages show narrative/style inconsistency
\end{itemize}
\vspace{2pt}

\textit{JSON Output Format:}

\begin{verbatim}
{
  "perspective_confusions": [{
    "fact_quote": "I could see the fear in Sarah's eyes...",
    "location": "Chapter 3, paragraph 4",
    "contradiction_pair": "He wondered what Sarah was thinking...",
    "contradiction_location": "Chapter 3, paragraph 4 (same paragraph)",
    "error_element": "point of view shift",
    "error_category": "perspective_confusion",
    "context": "Narrative perspective shifts from first-person..."
  }],
  "tone_inconsistencies": [{
    "fact_quote": "The funeral was a solemn affair...",
    "location": "Chapter 5, paragraph 2",
    "contradiction_pair": "Suddenly, Bob slipped on a banana peel...",
    "contradiction_location": "Chapter 5, paragraph 3",
    "error_element": "dramatic tone shift",
    "error_category": "tone_inconsistency",
    "context": "Serious funeral scene immediately followed..."
  }],
  "style_shifts": [{...}]
}
\end{verbatim}

\end{tcolorbox}
\caption{Complete judge prompt for Narrative \& Style category in \textsc{ConStory-Checker} evaluation protocol.}
\label{fig:storyverify-narrative}
\end{figure*}

\begin{figure*}[htbp]
\centering
\begin{tcolorbox}[width=0.95\textwidth, title=\textbf{Example 1: Timeline \& Plot Logic}, colback=pink!15, colframe=pink!60!red]
\scriptsize
\setlength{\parskip}{2pt}

\textbf{Error Element:} donation date \quad \textbf{Error Category:} absolute\_time\_error

\textbf{Fact Quote:} ``the box … bore a label scrawled in her grandmother's looping script: *Donations -- 1983.*''

\textbf{Location:} Chapter One, paragraph 2

\textbf{Contradiction Pair:} ``Your grandmother donated them when I was in middle school.''

\textbf{Contradiction Location:} Chapter One, paragraph 14

\textbf{Context:} The box is explicitly labeled as donated in 1983, yet Caroline recalls that her grandmother gave the books away when she was in middle school in the late 1970s.
\end{tcolorbox}

\vspace{0.2cm}

\begin{tcolorbox}[width=0.95\textwidth, title=\textbf{Example 2: Characterization}, colback=pink!15, colframe=pink!60!red]
\scriptsize
\setlength{\parskip}{2pt}

\textbf{Error Element:} hitchhiking duration \quad \textbf{Error Category:} memory\_contradiction

\textbf{Fact Quote:} ``I've been hitching rides across this sprawling country since I was twenty-one, a full decade before the millennium turned.''

\textbf{Location:} Chapter 1, paragraph 1

\textbf{Contradiction Pair:} ``About six years,'' I said.

\textbf{Contradiction Location:} Chapter 2, paragraph 11

\textbf{Context:} The narrator first claims to have been hitchhiking for a full decade, then later states it has been about six years.
\end{tcolorbox}

\vspace{0.2cm}

\begin{tcolorbox}[width=0.95\textwidth, title=\textbf{Example 3: World-building \& Setting}, colback=pink!15, colframe=pink!60!red]
\scriptsize
\setlength{\parskip}{2pt}

\textbf{Error Element:} operahouse status \quad \textbf{Error Category:} geographical\_contradiction

\textbf{Fact Quote:} ``In the stillness of my underground lair, carved beneath the crumbling ruins of a forgotten opera house, I play.''

\textbf{Location:} Paragraph 2

\textbf{Contradiction Pair:} ``It happened on a night when the opera house was alive with anticipation.''

\textbf{Contradiction Location:} Paragraph 11

\textbf{Context:} The opera house is described as a 'forgotten' ruin and yet later depicted as active and full of life.
\end{tcolorbox}

\vspace{0.2cm}

\begin{tcolorbox}[width=0.95\textwidth, title=\textbf{Example 4: Factual \& Detail Consistency}, colback=pink!15, colframe=pink!60!red]
\scriptsize
\setlength{\parskip}{2pt}

\textbf{Error Element:} character eye color \quad \textbf{Error Category:} appearance\_mismatch

\textbf{Fact Quote:} ``Dorian's jaw tightened, his blue eyes blazing with barely restrained fury.''

\textbf{Location:} Paragraph 6

\textbf{Contradiction Pair:} ``Kael met his twin's gaze, their matching emerald eyes locked in a silent battle of wills.''

\textbf{Contradiction Location:} Paragraph 17

\textbf{Context:} Dorian is first described as having blue eyes and later his eyes are described as emerald, a direct contradiction in the character's eye color.
\end{tcolorbox}

\vspace{0.2cm}

\begin{tcolorbox}[width=0.95\textwidth, title=\textbf{Example 5: Narrative \& Style}, colback=pink!15, colframe=pink!60!red]
\scriptsize
\setlength{\parskip}{2pt}

\textbf{Error Element:} point of view shift \quad \textbf{Error Category:} perspective\_confusion

\textbf{Fact Quote:} ``I stood frozen, my own breath hitching in my throat, watching the man I had called my closest friend collapse under the truth I had carried for so long.''

\textbf{Location:} The Weight of Truth, paragraph 1

\textbf{Contradiction Pair:} ``The cabin trembled under the weight of Froshi's grief.''

\textbf{Contradiction Location:} The Weight of Truth, paragraph 1

\textbf{Context:} Shifts from third-person narrative describing the cabin and Froshi's grief to first-person perspective without transition within the same paragraph.
\end{tcolorbox}

\caption{Representative error detection examples by \textsc{ConStory-Checker} across five consistency dimensions: Timeline \& Plot Logic, Characterization, World-building \& Setting, Factual \& Detail Consistency, and Narrative \& Style. Each example demonstrates the system's ability to identify subtle contradictions through structured extraction of conflicting passages with precise location references.}
\label{fig:storyverify-examples}
\end{figure*}


\FloatBarrier

\clearpage
\section{Additional Evaluation Results}
\label{sec:additional-eval}

This section provides supplementary visualization and analysis supporting the experimental findings presented in Section~\ref{sec:evaluation}.

\subsection{Model Performance Leaderboard}
\label{sec:leaderboard}

\begin{figure*}[!t]
\centering
\includegraphics[width=\textwidth]{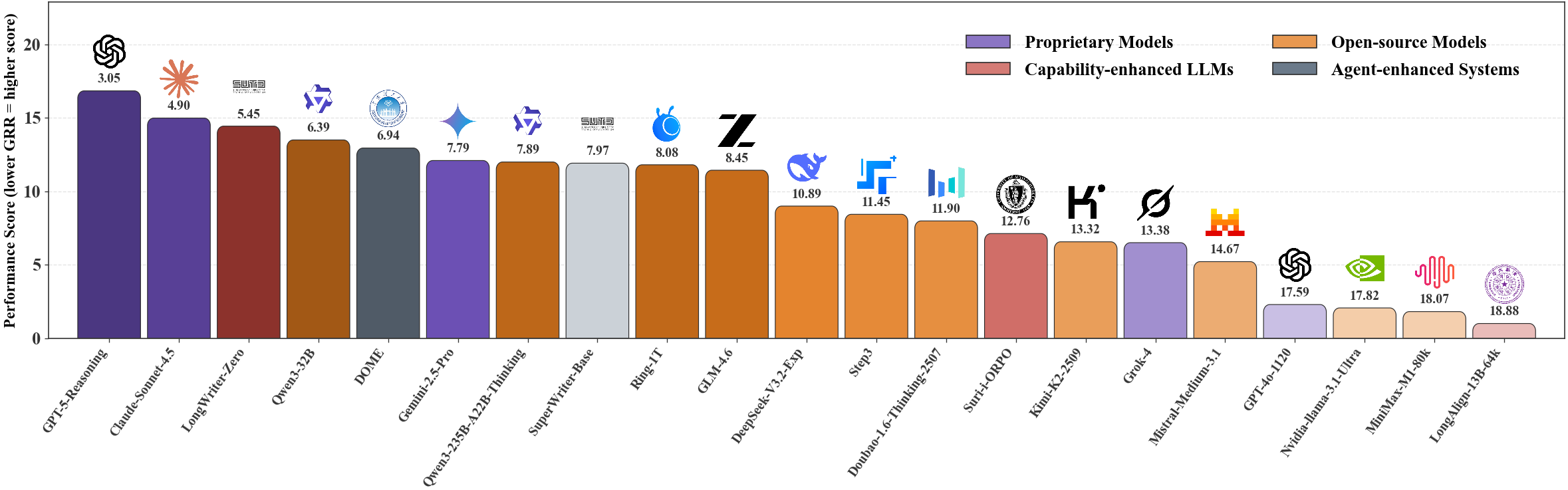}
\caption{Performance leaderboard of evaluated models based on GRR scores. Bar length indicates relative performance (longer bars represent better consistency), with color intensity reflecting score magnitude within each model category. Models are grouped by family: proprietary (top), open-source, capability-enhanced LLMs, and agent-enhanced systems (bottom).}
\label{fig:leaderboard}
\end{figure*}

\begin{figure*}[!t]
\centering
\includegraphics[width=\textwidth]{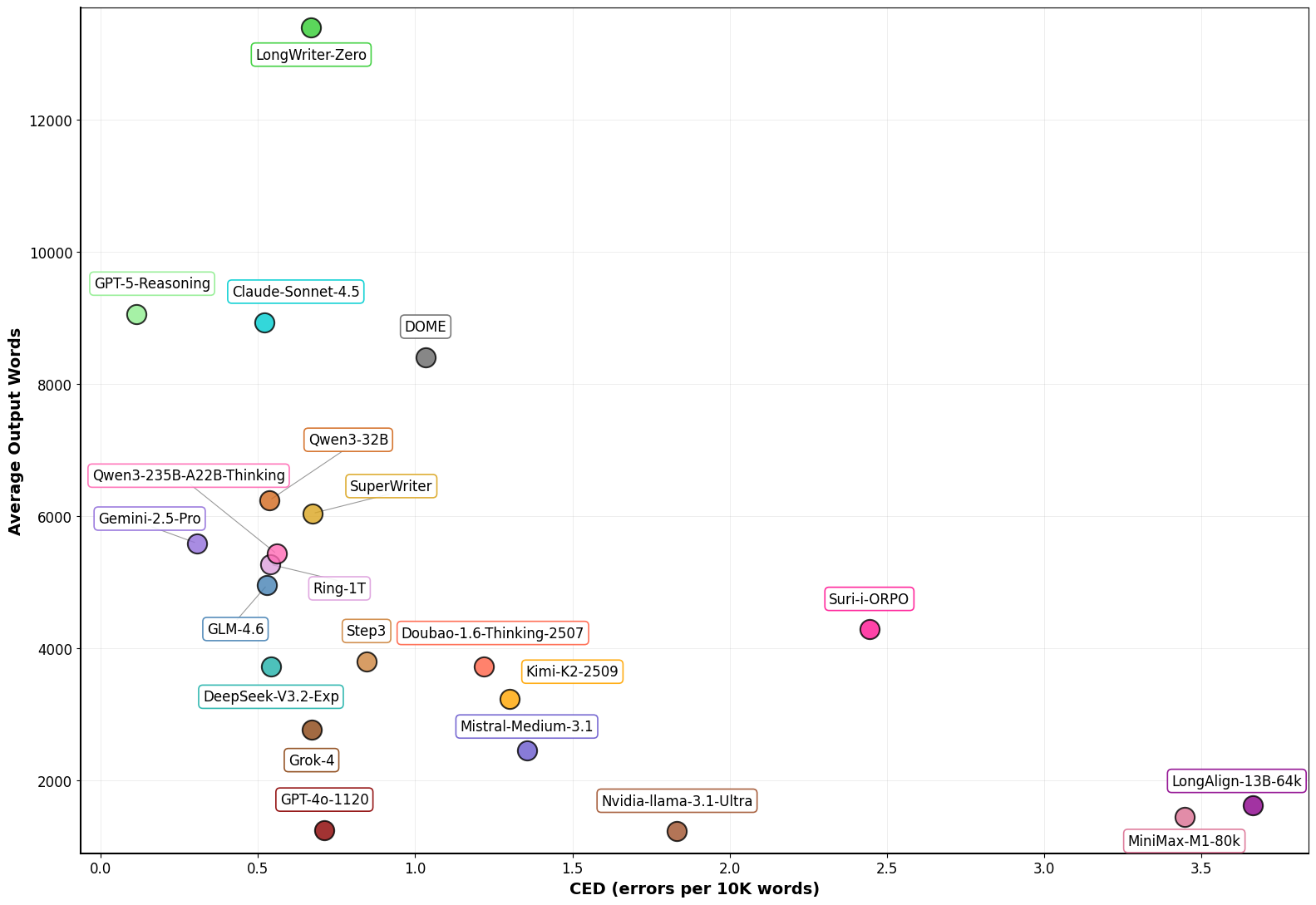}
\caption{The consistency performance (CED) of evaluated models versus average output words.}
\label{fig:scatter-ced-length}
\end{figure*}

To facilitate intuitive comparison of model consistency performance across different families, Figure~\ref{fig:leaderboard} presents a comprehensive leaderboard visualization based on the Group Relative Rank (GRR) metric introduced in Section~\ref{sec:rq1-benchmarks}. This visualization complements the quantitative results reported in Table~\ref{tab:comprehensive-performance} by providing a visual ranking that emphasizes relative performance differences across model families. The visualization employs an inverted transformation of GRR values, where each model's performance score is computed as $\text{score} = \max(\text{GRR}) - \text{GRR}_{\text{current}} + 1$. Since lower GRR values indicate superior performance, this transformation ensures that models with smaller GRR values receive higher scores and correspondingly longer bars, providing an intuitive visual representation where bar height directly correlates with model superiority. Within each category---proprietary models, open-source models, capability-enhanced LLMs, and agent-enhanced systems---color intensity maps to relative score magnitude, with darker shades representing higher performance and lighter shades indicating lower performance. This gradient encoding enables rapid visual identification of top performers within each model family while maintaining clear cross-category comparisons. Figure~\ref{fig:scatter-ced-length} further illustrates the relationship between model consistency performance (CED) and average output length.

Table~\ref{tab:task-type-ced} further disaggregates consistency performance by prompt task type, reporting CED scores across the four task categories defined in Section~\ref{sec:rq1-benchmarks}: Generation (748 prompts), Continuation (429 prompts), Expansion (419 prompts), and Completion (394 prompts). \textbf{Notably, Generation tasks consistently yield higher CED than other task types across most models, suggesting that open-ended story creation without prior context poses the greatest consistency challenge.}

\begin{table*}[t!]
\centering
\small
\caption{Consistency Error Density (CED) disaggregated by prompt task type. \textbf{CED}: errors per 10K words (lower is better). Task type columns show CED for each category: \textbf{Generation} (open-ended story creation), \textbf{Continuation} (extending provided narratives), \textbf{Expansion} (elaborating specific segments), and \textbf{Completion} (filling removed spans). Numbers in parentheses denote the number of prompts per task type. \textbf{Avg Words}: average output length. \textbf{Total}: number of completed stories. Models are grouped by family and sorted by Overall CED in ascending order within each category. \textbf{Bold values} indicate the highest CED among the four task types for each model.}
\label{tab:task-type-ced}
\setlength{\tabcolsep}{3.5pt}
\renewcommand{\arraystretch}{1.15}
\resizebox{\textwidth}{!}{
\begin{tabular}{l|c|c|c|c|c|c|c}
\toprule
\multirow{2}{*}{\textbf{Model}} & \multirow{2}{*}{\textbf{Overall CED}} & \textbf{Generation} & \textbf{Continuation} & \textbf{Expansion} & \textbf{Completion} & \multirow{2}{*}{\textbf{Avg Words}} & \multirow{2}{*}{\textbf{Total}} \\
& & \textbf{(748)} & \textbf{(429)} & \textbf{(419)} & \textbf{(394)} & & \\
\midrule
\multicolumn{8}{c}{\cellcolor[gray]{0.9}\textit{Proprietary Models}} \\
\midrule
GPT-5-Reasoning & 0.113 & 0.11 & 0.093 & 0.07 & \textbf{0.188} & 9050 & 1990 \\
Gemini-2.5-Pro & 0.302 & \textbf{0.379} & 0.233 & 0.277 & 0.257 & 5091 & 1996 \\
Gemini-2.5-Flash & 0.305 & 0.334 & 0.243 & 0.254 & \textbf{0.373} & 5504 & 1996 \\
Claude-Sonnet-4.5 & 0.52 & \textbf{0.67} & 0.387 & 0.498 & 0.402 & 8929 & 1998 \\
Grok-4 & 0.67 & \textbf{0.765} & 0.638 & 0.552 & 0.649 & 2765 & 2000 \\
GPT-4o-1120 & 0.711 & 0.776 & 0.389 & \textbf{0.912} & 0.708 & 1241 & 1774 \\
Doubao-1.6-Thinking-2507 & 1.217 & \textbf{1.415} & 1.154 & 1.084 & 1.054 & 3713 & 2000 \\
Mistral-Medium-3.1 & 1.355 & 1.376 & 0.931 & \textbf{2.02} & 1.069 & 2447 & 2000 \\
\midrule
\multicolumn{8}{c}{\cellcolor[gray]{0.9}\textit{Open-source Models}} \\
\midrule
GLM-4.6 & 0.528 & \textbf{0.785} & 0.311 & 0.381 & 0.437 & 4949 & 2000 \\
Qwen3-32B & 0.537 & \textbf{0.694} & 0.381 & 0.425 & 0.53 & 6237 & 2000 \\
Ring-1T & 0.539 & \textbf{0.641} & 0.484 & 0.489 & 0.461 & 5264 & 1999 \\
DeepSeek-V3.2-Exp & 0.541 & \textbf{0.795} & 0.325 & 0.382 & 0.465 & 3724 & 2000 \\
Qwen3-235B-A22B-Thinking & 0.559 & \textbf{0.605} & 0.44 & 0.575 & 0.586 & 5424 & 2000 \\
GLM-4.5 & 0.595 & 0.584 & 0.522 & \textbf{0.653} & 0.635 & 5421 & 2000 \\
Ling-1T & 0.699 & 0.72 & 0.597 & 0.613 & \textbf{0.862} & 5088 & 2000 \\
Step3 & 0.845 & 0.706 & 0.76 & 0.979 & \textbf{1.054} & 3793 & 1916 \\
Qwen3-Next-80B-Thinking & 0.959 & \textbf{1.15} & 0.913 & 0.778 & 0.846 & 4820 & 1973 \\
Kimi-K2-2509 & 1.3 & \textbf{1.686} & 0.926 & 1.162 & 1.112 & 3227 & 1792 \\
Kimi-K2-2507 & 1.33 & \textbf{1.775} & 0.933 & 1.109 & 1.152 & 3046 & 2000 \\
Qwen3-235B-A22B & 1.447 & 1.57 & 1.152 & \textbf{1.587} & 1.389 & 3246 & 2000 \\
Qwen3-Next-80B & 1.603 & \textbf{1.849} & 1.271 & 1.612 & 1.486 & 4013 & 2000 \\
Qwen3-4B-Instruct-2507 & 1.685 & 1.637 & 1.668 & \textbf{1.885} & 1.584 & 4919 & 1997 \\
Nvidia-llama-3.1-Ultra & 1.833 & \textbf{2.932} & 1.135 & 1.227 & 1.161 & 1224 & 1998 \\
Qwen3-30B-A3B-Instruct-2507 & 2.13 & \textbf{2.58} & 1.8 & 2.103 & 1.666 & 2968 & 2000 \\
DeepSeek-V3 & 2.422 & \textbf{3.18} & 2.102 & 2.001 & 1.781 & 670 & 2000 \\
QwenLong-L1-32B & 3.413 & \textbf{4.029} & 2.122 & 3.621 & 3.43 & 1234 & 2000 \\
DeepSeek-R1 & 3.419 & 3.007 & \textbf{3.829} & 3.737 & 3.415 & 680 & 1952 \\
MiniMax-M1-80k & 3.447 & 3.44 & 3.411 & \textbf{4.072} & 2.832 & 1442 & 1716 \\
\midrule
\multicolumn{8}{c}{\cellcolor[gray]{0.9}\textit{Capability-enhanced LLMs}} \\
\midrule
LongWriter-Zero-32B & 0.669 & \textbf{0.805} & 0.484 & 0.778 & 0.507 & 13393 & 1857 \\
Suri-i-ORPO & 2.445 & \textbf{2.768} & 2.117 & 2.355 & 2.284 & 4279 & 2000 \\
LongAlign-13B & 3.664 & \textbf{4.984} & 2.277 & 3.105 & 3.268 & 1624 & 2000 \\
\midrule
\multicolumn{8}{c}{\cellcolor[gray]{0.9}\textit{Agent-enhanced Systems}} \\
\midrule
SuperWriter & 0.674 & \textbf{0.75} & 0.632 & 0.673 & 0.576 & 6036 & 2000 \\
DOME & 1.033 & 1.108 & 0.912 & 0.94 & \textbf{1.122} & 8399 & 1969 \\
\bottomrule
\end{tabular}
}
\end{table*}


\subsection{Output Length Distribution Statistics}
\label{sec:length-statistics}

Table~\ref{tab:length-distribution-stats} provides detailed statistics on generated story lengths, complementing the analysis in Section~\ref{sec:rq2-length-dynamics}. For each model, we report the number and percentage of stories in five length categories (0--1k, 1k--3k, 3k--5k, 5k--8k, and 8k+ words), along with average word count and total completed stories.

These statistics show clear differences in generation strategies. Proprietary models like \textsc{GPT-5-Reasoning} and \textsc{Claude-Sonnet-4.5} prefer longer outputs (66.4\% and 66.8\% in the 8k+ category), while \textsc{GPT-4o-1120} and \textsc{Nvidia-llama-3.1-Ultra} generate mostly shorter texts below 3k words (85.0\% and 84.1\%). Open-source models such as \textsc{Qwen3-32B} and \textsc{GLM-4.6} show more balanced distributions across length bins. Combined with the error density metrics from Table~\ref{tab:comprehensive-performance}, these patterns highlight the trade-offs between generation length and consistency across different models.

\begin{table*}[t!]
\centering
\small
\caption{Detailed output length distribution across evaluated models. Columns show the number and percentage of stories in each word-count bin. Models are sorted by average word count in descending order.}
\label{tab:length-distribution-stats}
\setlength{\tabcolsep}{3.5pt}
\renewcommand{\arraystretch}{1.15}
\resizebox{\textwidth}{!}{
\begin{tabular}{l|c|c|c|c|c|c|c}
\toprule
\textbf{Model} & \textbf{0--1k} & \textbf{1k--3k} & \textbf{3k--5k} & \textbf{5k--8k} & \textbf{8k+} & \textbf{Avg Words} & \textbf{Total} \\
\midrule
\multicolumn{8}{c}{\cellcolor[gray]{0.9}\textit{Proprietary Models}} \\
\midrule
GPT-5-Reasoning & 24 (1.2\%) & 48 (2.0\%) & 58 (2.5\%) & 554 (27.8\%) & 1322 (66.4\%) & 9050 & 1990 \\
Claude-Sonnet-4.5 & 54 (2.7\%) & 40 (2.0\%) & 39 (2.0\%) & 531 (26.6\%) & 1334 (66.8\%) & 8929 & 1998 \\
Gemini-2.5-Flash & 48 (2.4\%) & 75 (3.8\%) & 830 (41.6\%) & 838 (42.0\%) & 205 (10.3\%) & 5504 & 1996 \\
Gemini-2.5-Pro & 43 (2.2\%) & 43 (2.2\%) & 845 (42.3\%) & 1024 (51.3\%) & 41 (2.1\%) & 5091 & 1996 \\
Doubao-1.6-Thinking-2507 & 40 (2.0\%) & 548 (27.4\%) & 1129 (56.5\%) & 255 (12.8\%) & 28 (1.4\%) & 3713 & 2000 \\
Grok-4 & 78 (3.9\%) & 1326 (66.3\%) & 542 (27.1\%) & 53 (2.6\%) & 1 (0.1\%) & 2765 & 2000 \\
Mistral-Medium-3.1 & 73 (3.6\%) & 1524 (76.2\%) & 380 (19.0\%) & 18 (0.9\%) & 5 (0.2\%) & 2447 & 2000 \\
GPT-4o-1120 & 196 (11.0\%) & 1578 (85.0\%) & 0 (0.0\%) & 0 (0.0\%) & 0 (0.0\%) & 1241 & 1774 \\
\midrule
\multicolumn{8}{c}{\cellcolor[gray]{0.9}\textit{Open-source Models}} \\
\midrule
Qwen3-32B & 48 (2.4\%) & 59 (2.9\%) & 21 (1.1\%) & 1872 (93.6\%) & 0 (0.0\%) & 6237 & 2000 \\
Qwen3-235B-A22B-Thinking & 60 (3.0\%) & 45 (2.2\%) & 174 (8.7\%) & 1721 (86.1\%) & 0 (0.0\%) & 5424 & 2000 \\
GLM-4.5 & 56 (2.8\%) & 177 (8.8\%) & 776 (38.8\%) & 698 (34.9\%) & 293 (14.6\%) & 5421 & 2000 \\
Ring-1T & 46 (2.3\%) & 48 (2.4\%) & 784 (39.2\%) & 1022 (51.1\%) & 99 (5.0\%) & 5264 & 1999 \\
Ling-1T & 45 (2.2\%) & 176 (8.8\%) & 892 (44.6\%) & 710 (35.5\%) & 177 (8.8\%) & 5088 & 2000 \\
GLM-4.6 & 49 (2.5\%) & 72 (3.6\%) & 953 (47.6\%) & 866 (43.3\%) & 60 (3.0\%) & 4949 & 2000 \\
Qwen3-4B-Instruct-2507 & 66 (3.3\%) & 35 (1.8\%) & 1188 (59.5\%) & 662 (33.1\%) & 46 (2.3\%) & 4919 & 1997 \\
Qwen3-Next-80B-Thinking & 80 (4.1\%) & 478 (24.2\%) & 951 (48.2\%) & 242 (12.3\%) & 222 (11.3\%) & 4828 & 1973 \\
Qwen3-Next-80B & 59 (2.9\%) & 114 (5.7\%) & 1632 (81.6\%) & 182 (9.1\%) & 13 (0.7\%) & 4013 & 2000 \\
Step3 & 45 (2.3\%) & 458 (23.9\%) & 1115 (58.2\%) & 272 (14.2\%) & 26 (1.4\%) & 3793 & 1916 \\
DeepSeek-V3.2-Exp & 50 (2.5\%) & 487 (24.3\%) & 1311 (65.5\%) & 227 (11.3\%) & 5 (0.2\%) & 3724 & 2000 \\
Qwen3-235B-A22B & 68 (3.4\%) & 353 (17.6\%) & 1576 (78.8\%) & 3 (0.1\%) & 0 (0.0\%) & 3246 & 2000 \\
Kimi-K2-2509 & 153 (8.5\%) & 771 (43.0\%) & 663 (37.0\%) & 138 (7.7\%) & 67 (3.7\%) & 3227 & 1792 \\
Kimi-K2-2507 & 69 (3.5\%) & 928 (46.4\%) & 948 (47.4\%) & 55 (2.8\%) & 0 (0.0\%) & 3046 & 2000 \\
Qwen3-30B-A3B-Instruct-2507 & 55 (2.8\%) & 948 (47.4\%) & 991 (49.5\%) & 1 (0.1\%) & 5 (0.2\%) & 2968 & 2000 \\
MiniMax-M1-80k & 694 (40.4\%) & 935 (54.5\%) & 11 (0.6\%) & 44 (2.6\%) & 32 (1.9\%) & 1442 & 1716 \\
DeepSeek-R1 & 108 (5.1\%) & 1852 (94.9\%) & 0 (0.0\%) & 0 (0.0\%) & 0 (0.0\%) & 1391 & 1952 \\
QwenLong-L1-32B & 792 (39.6\%) & 1188 (59.4\%) & 25 (1.2\%) & 2 (0.1\%) & 1 (0.1\%) & 1234 & 2000 \\
Nvidia-llama-3.1-Ultra & 317 (15.9\%) & 1681 (84.1\%) & 0 (0.0\%) & 0 (0.0\%) & 0 (0.0\%) & 1224 & 1998 \\
DeepSeek-V3 & 1971 (98.6\%) & 29 (1.5\%) & 0 (0.0\%) & 0 (0.0\%) & 0 (0.0\%) & 678 & 2000 \\
\midrule
\multicolumn{8}{c}{\cellcolor[gray]{0.9}\textit{Capability-enhanced LLMs}} \\
\midrule
LongWriter-Zero-32B & 58 (2.9\%) & 312 (15.7\%) & 188 (9.5\%) & 79 (4.0\%) & 1350 (67.9\%) & 13241 & 1987 \\
Suri-i-ORPO & 170 (8.5\%) & 840 (42.0\%) & 418 (20.9\%) & 228 (11.4\%) & 344 (17.2\%) & 4279 & 2000 \\
LongAlign-13B & 1812 (90.6\%) & 69 (3.5\%) & 0 (0.0\%) & 1 (0.1\%) & 118 (5.9\%) & 1624 & 2000 \\
\midrule
\multicolumn{8}{c}{\cellcolor[gray]{0.9}\textit{Agent-enhanced Systems}} \\
\midrule
DOME & 2 (0.1\%) & 4 (0.2\%) & 81 (4.1\%) & 536 (27.2\%) & 1346 (68.4\%) & 8399 & 1969 \\
SuperWriter & 59 (2.9\%) & 144 (7.2\%) & 378 (18.9\%) & 1069 (53.4\%) & 350 (17.5\%) & 6036 & 2000 \\
\bottomrule
\end{tabular}
}
\end{table*}


\subsection{Token-Level Uncertainty Metrics}
\label{sec:token-uncertainty}

Extending the entropy analysis presented in Section~\ref{sec:rq3-entropy}, this subsection provides comprehensive token-level uncertainty measurements across three complementary metrics: Shannon entropy, token probability, and perplexity. While Section~\ref{sec:rq3-entropy} demonstrated that error-bearing segments exhibit higher entropy than the whole-text baseline, a multi-metric analysis offers deeper insights into the probabilistic characteristics underlying consistency failures.

\paragraph{Metric Definitions.}
For each token position $t$ in a generated sequence, we compute three uncertainty measures from the model's output distribution. \textbf{Shannon Entropy} is defined in Section~\ref{sec:rq3-entropy}. \textbf{Token Probability} measures the model's confidence in the selected token $w_t$, computed as $p_t = \exp(\log p(w_t | w_{<t}))$, where higher values indicate stronger confidence. \textbf{Perplexity} captures the model's surprise at the observed token sequence, calculated as the exponential of average negative log-probability:
\begin{equation}
\text{PPL}(S) = \exp\left(-\frac{1}{N} \sum_{t=1}^{N} \log p(w_t | w_{<t})\right),
\end{equation}
with lower perplexity indicating more predictable sequences. For text segments $S$ with $N$ tokens, we report segment-level averages: $\bar{p}(S) = \frac{1}{N}\sum_{t=1}^{N} p_t$, and $\overline{\text{PPL}}(S) = \frac{1}{N}\sum_{t=1}^{N} \frac{1}{p_t}$.

\paragraph{Results.}
Table~\ref{tab:token-uncertainty-comparison} presents comprehensive comparisons across all three metrics for two representative models. The results reveal consistent patterns: error content consistently exhibits \textbf{higher uncertainty} (elevated entropy and perplexity) and \textbf{lower confidence} (reduced probability) compared to the whole-text baseline. For \textsc{Qwen3-30B-A3B-Instruct-2507}, error segments show \textbf{+12.03\%} higher entropy, \textbf{-5.41\%} lower probability, and \textbf{+2.54\%} higher perplexity relative to whole text, while \textsc{Qwen3-4B-Instruct-2507} demonstrates even stronger divergence (\textbf{+19.24\%}, \textbf{-7.99\%}, \textbf{+5.55\%} respectively). These converging signals across all three metrics indicate that \textbf{consistency failures emerge precisely in regions where the model exhibits elevated uncertainty and diminished confidence}, suggesting that token-level uncertainty provides a reliable early warning signal for potential narrative inconsistencies during generation.

\begin{table*}[t!]
\centering
\small
\caption{Comprehensive token-level uncertainty comparison across three metrics. Each metric compares error-bearing segments against the whole-text baseline. Higher entropy and perplexity, along with lower probability, indicate greater model uncertainty. Relative differences show the percentage change of error content compared to whole text.}
\label{tab:token-uncertainty-comparison}
\setlength{\tabcolsep}{6pt}
\renewcommand{\arraystretch}{1.2}
\begin{tabular}{l|cc|c}
\toprule
\multirow{2}{*}{\textbf{Model}} & \multicolumn{2}{c|}{\textbf{Average Values}} & \textbf{Relative Difference} \\
\cmidrule(lr){2-3} \cmidrule(lr){4-4}
& \textbf{Whole Text} & \textbf{Error Content} & \textbf{(Error vs Whole)} \\
\midrule
\multicolumn{4}{c}{\cellcolor[gray]{0.9}\textit{Entropy (bits) --- Higher indicates greater uncertainty}} \\
\midrule
Qwen3-30B-A3B-Instruct-2507 & 1.1438 & 1.2814 & \textbf{\textcolor{red!70!black}{+12.03\%}} \\
Qwen3-4B-Instruct-2507 & 1.0734 & 1.2799 & \textbf{\textcolor{red!70!black}{+19.24\%}} \\
\midrule
\multicolumn{4}{c}{\cellcolor[gray]{0.9}\textit{Probability --- Higher indicates greater confidence}} \\
\midrule
Qwen3-30B-A3B-Instruct-2507 & 0.6895 & 0.6522 & \textbf{\textcolor{green!50!black}{-5.41\%}} \\
Qwen3-4B-Instruct-2507 & 0.7097 & 0.6530 & \textbf{\textcolor{green!50!black}{-7.99\%}} \\
\midrule
\multicolumn{4}{c}{\cellcolor[gray]{0.9}\textit{Perplexity --- Lower indicates better predictability}} \\
\midrule
Qwen3-30B-A3B-Instruct-2507 & 1.8875 & 1.9354 & \textbf{\textcolor{red!70!black}{+2.54\%}} \\
Qwen3-4B-Instruct-2507 & 1.8566 & 1.9596 & \textbf{\textcolor{red!70!black}{+5.55\%}} \\
\bottomrule
\end{tabular}
\end{table*}


\FloatBarrier
\clearpage
\subsection{Model-Specific Error Correlations}
\label{sec:model-error-correlations}

\begin{figure*}[t!]
\centering
\includegraphics[width=\textwidth]{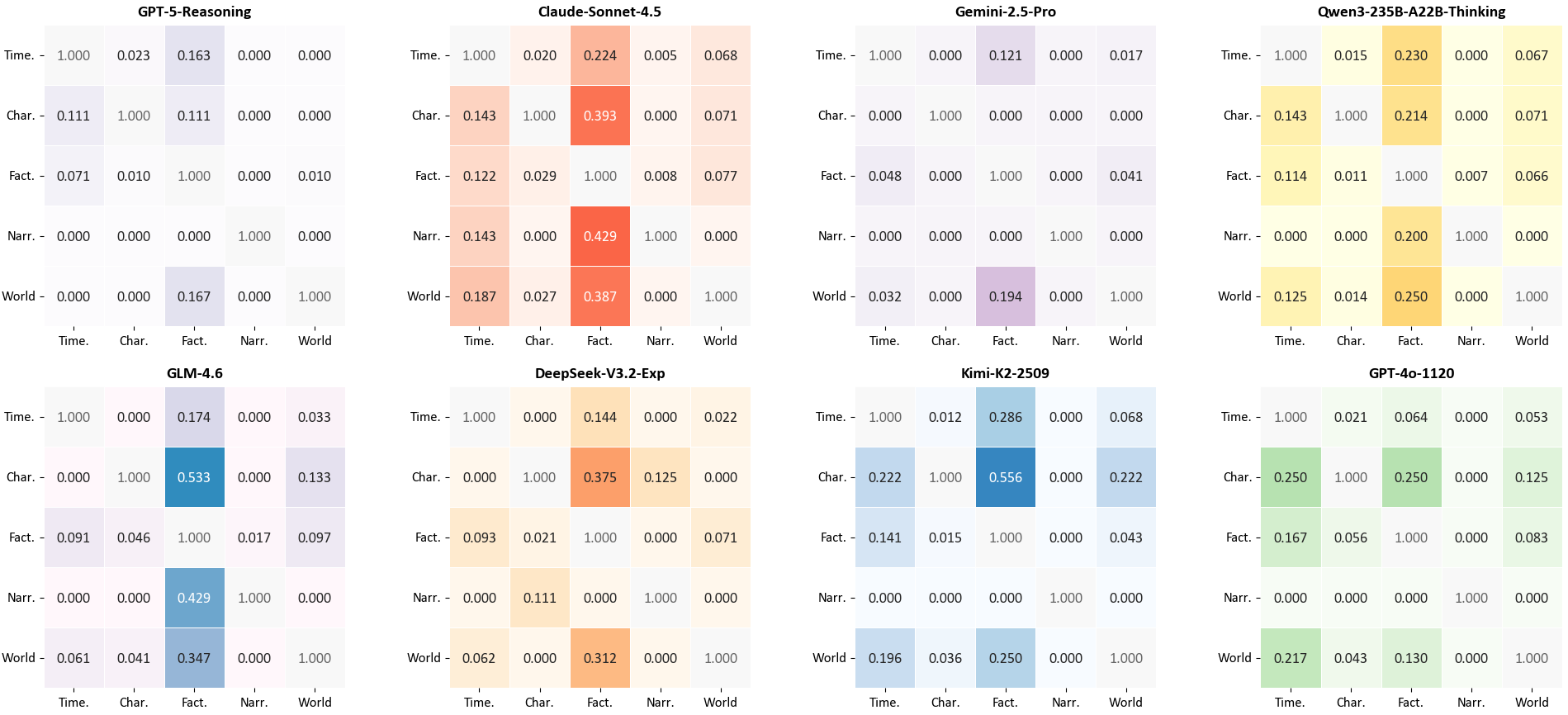}
\caption{Model-specific error correlation matrices across eight representative models. Darker colors indicate stronger positive correlations between error categories.}
\label{fig:model-error-correlations}
\end{figure*}

Figure~\ref{fig:model-error-correlations} shows model-specific correlation matrices for eight representative models, extending the analysis in Section~\ref{sec:rq4-correlations}. Proprietary models (\textsc{GPT-5-Reasoning}, \textsc{Gemini-2.5-Pro}) have sparse matrices with weak cross-category dependencies, while \textsc{Claude-Sonnet-4.5} shows stronger Fact.--World (r=0.387) and Narr.--Fact. (r=0.429) correlations. Among open-source models, \textsc{GLM-4.6} and \textsc{Kimi-K2-2509} show the strongest Char.--Fact. correlations (r=0.533 and r=0.556, respectively).

\subsection{Extended Positional Analysis}
\label{sec:extended-positional}

Table~\ref{tab:extended-error-position} extends the positional analysis presented in Section~\ref{sec:rq5-positions} by providing per-model statistics for eight representative models. The extended analysis confirms that the positional patterns observed in the main text---fact positions clustering in the early-to-mid narrative (15--30\%) and contradiction positions extending toward later sections (40--60\%)---hold consistently across diverse model architectures. In the table, \colorbox[rgb]{.741,.843,.933}{\textbf{blue}} highlights the largest Gap values per error type.

\begin{table*}[t!]
\centering
\small
\setlength{\tabcolsep}{3pt}
\renewcommand{\arraystretch}{1.15}
\caption{Per-model positional distribution of seven representative error subtypes across eight models. Positions are normalized by story length (0--100\%). \textbf{Fact}: average position where facts are first established; \textbf{Contra}: average position where contradictions appear; \textbf{Gap}: average distance between fact and contradiction positions, computed as $\text{Avg Gap} = \frac{1}{n}\sum_{i=1}^{n}|\text{contra}_i - \text{fact}_i|$. \colorbox[rgb]{.741,.843,.933}{\textbf{Blue}} = largest Gap per error type.}
\label{tab:extended-error-position}
\resizebox{\textwidth}{!}{
\begin{tabular}{ll ccccccc}
\toprule
\multirow{2}{*}{\textbf{Model}} & \multirow{2}{*}{\textbf{Metric}} & \textbf{Absolute Time} & \textbf{Core Rules} & \textbf{Quantitative} & \textbf{Geographical} & \textbf{Nomenclature} & \textbf{Memory} & \textbf{Perspective} \\
& & \textbf{Contradictions} & \textbf{Violations} & \textbf{Mismatches} & \textbf{Contradictions} & \textbf{Confusions} & \textbf{Contradictions} & \textbf{Confusions} \\
\midrule
\rowcolor[gray]{0.95}
\multicolumn{9}{c}{\textit{Proprietary Models}} \\
\midrule
\multirow{3}{*}{GPT-5-Reasoning}
& Fact   & 48.4\% & 21.8\% & 20.0\% & 7.7\% & 20.2\% & 21.4\% & 13.3\% \\
& Contra & 90.6\% & 21.1\% & 32.2\% & 30.2\% & 38.1\% & 40.5\% & 4.3\% \\
& Gap    & \cellcolor[rgb]{.741,.843,.933}\textbf{47.7\%} & 0.8\% & 12.7\% & 22.6\% & 24.2\% & 19.1\% & \cellcolor[rgb]{.741,.843,.933}\textbf{15.8\%} \\
\midrule
\multirow{3}{*}{Claude-Sonnet-4.5}
& Fact   & 23.1\% & 19.1\% & 25.7\% & 12.2\% & 18.6\% & 16.3\% & 35.7\% \\
& Contra & 41.5\% & 43.1\% & 43.4\% & 32.0\% & 37.7\% & 34.5\% & 35.9\% \\
& Gap    & 33.3\% & 28.1\% & 25.0\% & 21.3\% & \cellcolor[rgb]{.741,.843,.933}\textbf{29.1\%} & 19.7\% & 1.0\% \\
\midrule
\multirow{3}{*}{Gemini-2.5-Pro}
& Fact   & 19.3\% & 12.9\% & 21.1\% & 14.4\% & 17.7\% & 28.7\% & 0.2\% \\
& Contra & 38.3\% & 39.8\% & 44.5\% & 41.9\% & 30.1\% & 29.5\% & 1.7\% \\
& Gap    & 19.0\% & 27.0\% & \cellcolor[rgb]{.741,.843,.933}\textbf{28.6\%} & 28.2\% & 18.8\% & 1.9\% & 1.5\% \\
\midrule
\multirow{3}{*}{GPT-4o-1120}
& Fact   & 13.7\% & 42.9\% & 33.9\% & 31.7\% & 11.0\% & 13.7\% & 23.9\% \\
& Contra & 52.9\% & 55.8\% & 41.0\% & 51.2\% & 30.2\% & 46.3\% & 15.5\% \\
& Gap    & 39.2\% & \cellcolor[rgb]{.741,.843,.933}\textbf{37.6\%} & 26.8\% & \cellcolor[rgb]{.741,.843,.933}\textbf{44.5\%} & 19.2\% & 32.5\% & 11.4\% \\
\midrule
\rowcolor[gray]{0.95}
\multicolumn{9}{c}{\textit{Open-source Models}} \\
\midrule
\multirow{3}{*}{\shortstack[l]{Qwen3-235B\\-A22B-Thinking}}
& Fact   & 16.4\% & 24.6\% & 18.5\% & 19.2\% & 19.4\% & 12.5\% & 28.4\% \\
& Contra & 36.6\% & 37.0\% & 35.6\% & 34.9\% & 37.3\% & 32.0\% & 27.4\% \\
& Gap    & 20.2\% & 23.5\% & 21.7\% & 19.4\% & 23.5\% & 26.3\% & 3.3\% \\
\midrule
\multirow{3}{*}{GLM-4.6}
& Fact   & 17.7\% & 16.6\% & 21.0\% & 24.9\% & 22.3\% & 29.4\% & 0.1\% \\
& Contra & 45.7\% & 36.8\% & 40.6\% & 45.3\% & 32.8\% & 33.7\% & 1.3\% \\
& Gap    & 28.0\% & 26.6\% & 25.5\% & 41.7\% & 27.0\% & 22.2\% & 1.2\% \\
\midrule
\multirow{3}{*}{DeepSeek-V3.2-Exp}
& Fact   & 15.1\% & 25.0\% & 23.8\% & 30.7\% & 25.3\% & 29.3\% & 0.4\% \\
& Contra & 36.3\% & 42.5\% & 41.3\% & 31.7\% & 31.1\% & 46.6\% & 1.7\% \\
& Gap    & 21.8\% & 17.5\% & 23.7\% & 37.4\% & 21.4\% & \cellcolor[rgb]{.741,.843,.933}\textbf{56.4\%} & 1.4\% \\
\midrule
\multirow{3}{*}{Kimi-K2-2509}
& Fact   & 26.8\% & 26.6\% & 23.5\% & 22.3\% & 38.1\% & 23.1\% & 7.4\% \\
& Contra & 49.3\% & 38.7\% & 46.3\% & 46.1\% & 38.4\% & 42.4\% & 9.6\% \\
& Gap    & 28.3\% & 26.5\% & 26.6\% & 33.2\% & 23.5\% & 25.0\% & 2.4\% \\
\bottomrule
\end{tabular}
} 
\end{table*}



\clearpage
\section{Explanation of Metrics}
\label{sec:metrics-explanation}

\subsection{Example Calculation}
\label{sec:example-calculation}

We illustrate CED and GRR computation with examples demonstrating their complementary roles.

\paragraph{CED Calculation.}
Consider models generating stories:

\begin{table}[h]
\centering
\small
\setlength{\tabcolsep}{10pt}
\begin{tabular}{lrr}
\toprule
& \textbf{Words} & \textbf{Errors} \\
\midrule
Story 1 & 8,000 & 2 \\
Story 2 & 10,000 & 3 \\
Story 3 & 6,000 & 1 \\
Story 4 & 8,000 & 0 \\
Story 5 & 800 & 0 \\
\midrule
\textbf{Total (1--5)} & \textbf{32,800} & \textbf{6} \\
\bottomrule
\end{tabular}
\end{table}

For Stories 1--5, overall CED normalizes total errors by total words (per 10K):
\[
\text{CED}_{\text{overall}} = \frac{6}{32{,}800 / 10{,}000} = \frac{6}{3.28} \approx 1.83
\]

Category CED: If errors are 1 \textit{Char.}, 1 \textit{Fact.}, 1 \textit{Narr.}, 2 \textit{Time.}, 1 \textit{World}:
\[
\text{CED}_{\text{Time}} = \frac{2}{3.28} \approx 0.61, \quad \text{CED}_{\text{Char}} = \frac{1}{3.28} \approx 0.30
\]

However, Stories 4 and 5 both have zero errors, yielding identical CED=0.00, yet Story 4 generates 8,000 words while Story 5 generates only 800 words—a 10-fold difference in narrative completeness that CED cannot capture.

\paragraph{GRR Calculation.}
To address this, GRR ranks models within each story using the quality score from Equation~\eqref{eq:quality-rq1}. For Stories 4 and 5:
\[
Q_{4} = \frac{8{,}000}{1+0} = 8{,}000, \quad Q_{5} = \frac{800}{1+0} = 800
\]

Story 4 ranks higher (rank 1) than Story 5 (rank 2) despite identical CED. GRR then averages these ranks across all stories following Equation~\eqref{eq:grr-rq1}, where lower values indicate better performance.

\paragraph{Interpretation.}
In Table~\ref{tab:comprehensive-performance}: \textbf{CED} reports absolute error density (errors per 10K words); \textbf{GRR} provides relative ranking that accounts for both consistency and completeness, addressing CED's inability to differentiate models when error densities are identical.





